\documentclass[acmsmall]{acmart}

\AtBeginDocument{%
  }

\setcopyright{cc}
\setcctype{by-nc-nd}
\acmDOI{10.1145/3798217}
\acmYear{2026}
\acmJournal{PACMPL}
\acmVolume{10}
\acmNumber{OOPSLA1}
\acmArticle{109}
\acmMonth{4}
\received{2025-10-10}
\received[accepted]{2026-02-17}

\usepackage{mathpartir}
\usepackage{graphicx} 
\usepackage{enumitem}
\usepackage{tightenum}
\usepackage{listings}
\usepackage{tabularx}
\usepackage{array} 
\usepackage{diagbox} 
\usepackage{setspace}
\usepackage{epsfig}
\usepackage{graphicx}

\usepackage{amsmath}
\usepackage{float}
\usepackage{capt-of}
\usepackage{fvextra}   
\usepackage{mdframed}  
\usepackage[most]{tcolorbox}

\usepackage{url}
\usepackage{subcaption}
\usepackage[ruled]{algorithm2e}
\usepackage{booktabs}
\usepackage{multirow}
\usepackage{diagbox}
\usepackage{textcomp}
\usepackage{placeins}
\usepackage{parcolumns}
\tcbuselibrary{listings}

\usepackage[T1]{fontenc}
\usepackage{tikz}
\usetikzlibrary{arrows.meta,positioning}

\newcommand{\shihao}[1]{{\color{orange} \sf (SH: #1)}}

\newcommand{\mengting}[1]{{\color{brown} \sf (MT: #1)}}
\newcommand{\songlh}[1]{{\color{red} \sf (LH: #1)}}

  \newcommand{\commentty}[1]{{\color{purple} \sf (TY: #1)}}

\newcommand{\ignore}[1]{}

\newcommand{\bolditalicparagraphnospace}[1]{\noindent{{\textit{\textbf{#1}}}}}

\newcommand{\bolditalicparagraph}[1]{\vspace*{0.5ex}\noindent{{\textit{\textbf{#1}}}}}

\newcommand{\italicparagraph}[1]{\noindent\underline{\textit{#1}}}
\newcommand{\indentitalicparagraph}[1]{\underline{\textit{#1}}}

\newcommand{\boldunderparagraph}[1]{\vspace*{0.5ex}\noindent\underline{\textbf{#1}}}
\newcommand{\indentboldunderparagraph}[1]{\vspace*{0.5ex}\underline{\textbf{#1}}}

\newcommand{\Tool}{SymGPT\xspace}

\newcommand{\beforecaption}{\vspace{-.15cm}\begin{spacing}{0.85}}
\newcommand{\aftercaption}{\vspace{-.15cm}\end{spacing}}

\newcommand{\mycaption}[3]{\beforecaption\caption{\label{#1}{#2} #3}\aftercaption}

\newcommand{\mysmallcaption}[3]{\beforecaption\caption{\footnotesize \label{#1}{\footnotesize #2} \footnotesize #3}\aftercaption}

\newcommand{\eg}{\textit{e.g.}}
\newcommand{\ie}{\textit{i.e.}}

\newcounter{insight}

\newcolumntype{Y}{>{\centering\arraybackslash}X}
\newcommand{\var}[1]{\mathit{#1}}

\definecolor{promptbg}{HTML}{F8F9FF}
\definecolor{promptborder}{HTML}{DDE3FF}

\newcommand{\hlline}[1]{%
  \rlap{%
    \color{yellow!50}%
    \rule[-0.3\baselineskip]{#1\linewidth}{\baselineskip}%
  }%
}


\begin{document}

\title{SymGPT: Auditing Smart Contracts via Combining Symbolic Execution with Large Language Models}

\author{Shihao Xia}
\orcid{0009-0006-7334-7701}
\affiliation{%
  \institution{The Pennsylvania State University}
  \city{State College}
  \country{USA}
}
\email{szx5097@psu.edu}

\author{Mengting He}
\orcid{0009-0006-5289-2364}
\affiliation{%
  \institution{The Pennsylvania State University}
  \city{State College}
  \country{USA}
}
\email{mvh6224@psu.edu}

\author{Shuai Shao}
\orcid{0000-0001-6736-8393}
\affiliation{%
  \institution{University of Connecticut}
  \city{Storrs}
  \country{USA}
}
\email{shuai.shao@uconn.edu}

\author{Tingting Yu}
\orcid{0000-0002-9461-4251}
\affiliation{%
  \institution{University of Connecticut}
  \city{Storrs}
  \country{USA}
}
\email{tingting.yu@uconn.edu}

\author{Yiying Zhang}
\orcid{0009-0005-6263-7802}
\affiliation{%
  \institution{University of California San Diego}
  \city{La Jolla}
  \country{USA}
}
\email{yiying@ucsd.edu}

\author{Nobuko Yoshida}
\orcid{0000-0002-3925-8557}
\affiliation{%
  \institution{University of Oxford}
  \city{Oxford}
  \country{United Kingdom}
}
\email{nobuko.yoshida@cs.ox.ac.uk}

\author{Linhai Song}
\authornote{Corresponding author.}
\orcid{0000-0002-3185-9278}
\affiliation{%
  \institution{SKLP, Institute of Computing Technology, Chinese Academy of Sciences}
  \city{Beijing}
  \country{China}
}
\email{songlinhai@ict.ac.cn}

\definecolor{sol-bg}{HTML}{F8FAFF}     
\definecolor{sol-frame}{HTML}{D8E1FF}  
\definecolor{sol-num}{HTML}{9AA4B2}    
\definecolor{sol-key}{HTML}{1D4ED8}    
\definecolor{sol-type}{HTML}{0B7A75}   
\definecolor{sol-str}{HTML}{8B5CF6}    
\definecolor{sol-cmt}{HTML}{6B7280}    
\definecolor{sol-lit}{HTML}{B45309}    
\definecolor{sol-id}{HTML}{111827}     

\lstdefinelanguage{Solidity}{
basicstyle=\ttfamily\scriptsize,
  stepnumber=1,
  numbersep=5pt,
  breaklines=true,
  showstringspaces=false,
  frame=single,
  tabsize=2,
  keywords=[1]{anonymous, assembly, assert, balance, break, call, callcode, case, catch, class, constant, continue, constructor, contract, debugger, default, delegatecall, delete, do, else, emit, event, experimental, export, external, false, finally, for, function, gas, if, implements, import, in, indexed, instanceof, interface, internal, is, length, library, log0, log1, log2, log3, log4, memory, modifier, new, payable, pragma, private, protected, public, pure, push, require, return, returns, revert, selfdestruct, send, solidity, storage, struct, suicide, super, switch, then, this, throw, transfer, true, try, typeof, using, view, while, with, addmod, ecrecover, keccak256, mulmod, ripemd160, sha256, sha3}, 
	keywordstyle=[1]\color{blue}\bfseries,
	keywords=[2]{address, bool, byte, bytes, bytes1, bytes2, bytes3, bytes4, bytes5, bytes6, bytes7, bytes8, bytes9, bytes10, bytes11, bytes12, bytes13, bytes14, bytes15, bytes16, bytes17, bytes18, bytes19, bytes20, bytes21, bytes22, bytes23, bytes24, bytes25, bytes26, bytes27, bytes28, bytes29, bytes30, bytes31, bytes32, enum, int, int8, int16, int24, int32, int40, int48, int56, int64, int72, int80, int88, int96, int104, int112, int120, int128, int136, int144, int152, int160, int168, int176, int184, int192, int200, int208, int216, int224, int232, int240, int248, int256, mapping, string, uint, uint8, uint16, uint24, uint32, uint40, uint48, uint56, uint64, uint72, uint80, uint88, uint96, uint104, uint112, uint120, uint128, uint136, uint144, uint152, uint160, uint168, uint176, uint184, uint192, uint200, uint208, uint216, uint224, uint232, uint240, uint248, uint256, var, void, ether, finney, szabo, wei, days, hours, minutes, seconds, weeks, years},	
	keywordstyle=[2]\color{teal}\bfseries,
	keywords=[3]{block, blockhash, coinbase, difficulty, gaslimit, number, timestamp, msg, data, gas, sender, sig, now, tx, gasprice, origin},	
	keywordstyle=[3]\color{violet}\bfseries,
	identifierstyle=\color{black},
	sensitive=true,
	comment=[l]{//},
	morecomment=[s]{/*}{*/},
	commentstyle=\color{gray}\ttfamily,
	stringstyle=\color{red}\ttfamily,
	morestring=[b]',
	morestring=[b]",
  morecomment=[f][\color{red}]{-\ }
}

\lstdefinestyle{solidity-gorgeous}{
  language=Solidity,
  basicstyle=\ttfamily\scriptsize,
  numbers=left,
  numberstyle=\tiny\color{sol-num},
  numbersep=10pt,
  stepnumber=1,
  showstringspaces=false,
  columns=fullflexible,
  keepspaces=true,
  upquote=true,
  tabsize=2,
  breaklines=true,
  breakatwhitespace=false,
  keywordstyle=\bfseries\color{sol-key},
  commentstyle=\itshape\color{sol-cmt},
  stringstyle=\color{sol-str},
  identifierstyle=\color{sol-id},
  literate=
    *{0}{{{\color{sol-lit}0}}}1
     {1}{{{\color{sol-lit}1}}}1
     {2}{{{\color{sol-lit}2}}}1
     {3}{{{\color{sol-lit}3}}}1
     {4}{{{\color{sol-lit}4}}}1
     {5}{{{\color{sol-lit}5}}}1
     {6}{{{\color{sol-lit}6}}}1
     {7}{{{\color{sol-lit}7}}}1
     {8}{{{\color{sol-lit}8}}}1
     {9}{{{\color{sol-lit}9}}}1,
  emph={address,uint,uint8,uint16,uint32,uint64,uint128,uint256,int,int256,
        string,bool,bytes,bytes4,bytes32,IERC721,IERC165},
  emphstyle=\color{sol-type}\bfseries,
}

\lstset{emph={msg.sender},emphstyle=\color{red}}

\lstdefinelanguage{comment}{
basicstyle=\ttfamily\scriptsize,
    numbersep = 5pt,
    breaklines = true,
    showstringspaces = false,
    breakindent = 0em,
    frame=single
}

\definecolor{phText}{HTML}{1E3A8A} 
\definecolor{phBg}{HTML}{DBEAFE}   
\newcommand{\PHkeep}[1]{\textcolor{phText}{\colorbox{phBg}{\{\{#1\}\}}}}

\lstdefinelanguage{GPTPrompt}{
basicstyle=\ttfamily\scriptsize,
  numbersep=5pt,
  breaklines=true,
  showstringspaces=false,
  breakindent=0em,
  frame=single,
  moredelim=[is][\PHkeep]{\{\{}{\}\}}
}

\lstdefinelanguage{YAML}{
basicstyle=\ttfamily\scriptsize,
  numbersep=5pt,
  breaklines=true,
  showstringspaces=false,
  breakindent=1em,
  frame=single,
    keywords={true,false,null,y,n},
    keywordstyle=\color{darkgray}\bfseries,
    basicstyle=\ttfamily\footnotesize,
    sensitive=false,
    comment=[l]{\#},
    morecomment=[s]{/*}{*/},
    commentstyle=\color{purple}\ttfamily,
    stringstyle=\color{red}\ttfamily,
    morestring=[b]',
    morestring=[b]"
}
\lstdefinelanguage{ebnf}%
{
linewidth=\columnwidth,
  basicstyle=\ttfamily,
  stepnumber=1,
  numbersep=5pt,
  breaklines=true,
  showstringspaces=false,
  tabsize=2,
  frame=single,
  keywords=[1]{if, then, with, and, or, not, msg, sender}, 
  keywordstyle=[1]\color{blue}\bfseries,
  keywords=[2]{check, throw, call, return, emit, assign, follow, b_exp, b_op,
  value, field, index, order, constant, change, mint, burn, field_name, para, key, anchor_fun, flag, fun, event, op, field_value},	
	keywordstyle=[2]\color{teal}\bfseries,
 keywords=[3]{checkThrow, checkCall, checkEmit, checkEndValue, checkOrder, getFieldValue, getField,  getPara, getOrder, checkChange, checkMint, checkBurn},	
	keywordstyle=[3]\color{violet}\bfseries,
  sensitive=true,
	comment=[l]{//},
	morecomment=[s]{/*}{*/},
	commentstyle=\color{gray}\ttfamily,
	stringstyle=\color{black}\ttfamily,
	morestring=[b]',
	morestring=[b]",
  morecomment=[f][\color{red}]{-\ },
}

\begin{abstract}
To govern smart contracts running on Ethereum, 
multiple Ethereum Request for Comment (ERC) standards have 
been developed, each defining a set of rules governing contract behavior. 
Violating these rules can cause 
serious security issues and financial losses, signifying the importance 
of verifying ERC compliance. Today's practices of such verification include manual audits, expert-developed program-analysis tools, and large language models (LLMs), all of which remain ineffective at detecting ERC rule violations.

This paper introduces \emph{\Tool{}}, 
a tool that combines LLMs with 
symbolic execution to automatically 
verify smart contracts' compliance with ERC rules. 
We begin by empirically analyzing 132 ERC rules from three major
ERC standards, examining their content, security implications, 
and natural language descriptions. Based on this study, 
\Tool{} instructs an LLM to translate ERC rules into a domain-specific language, 
synthesizes constraints from the translated rules to model potential rule violations,
and performs symbolic execution for violation detection. 
Our evaluation shows that \Tool{} identifies 5,783 ERC rule violations 
in 4,000 real-world contracts, including 1,375 violations with 
clear attack paths for financial theft. 
Furthermore, \Tool{} outperforms six automated techniques 
and a security-expert auditing service, underscoring its superiority over current 
smart contract analysis methods.

\end{abstract}


\maketitle

\enlargethispage{0.1in}
\section{Introduction}
\label{sec:intro}

\boldunderparagraph{Ethereum and ERC.}
Since Bitcoin's creation, blockchain technology has significantly evolved. A key development is Ethereum \cite{eth-1,eth-2}, a decentralized, open-source platform that enables the creation and execution of decentralized applications (DApps) and smart contracts—self-executing agreements written into code \cite{dapps, sc-anatomy}.
To govern smart contracts on Ethereum, formal standards known as Ethereum Request for Comments (\textit{ERCs}) have been established \cite{token-standard}. 
For example, the ERC20 standard defines the rules for fungible tokens~\cite{erc20}, 
whereas ERC721 sets the requirements for non-fungible tokens (NFTs)~\cite{erc721}.
ERCs are vital in the Ethereum ecosystem, providing a common set of specifications 
that ensure interoperability and compatibility among various Ethereum-based projects, wallets, and DApps~\cite{dapps}.

{
\setlength{\belowcaptionskip}{3pt}
\begin{figure}[t]

\begin{minipage}{\columnwidth}
\begin{center}
\begin{lstlisting}[
numbers=left, xleftmargin=.20in, framexleftmargin=.16in, framexrightmargin=-0.05in,language=Solidity,basicstyle=\ttfamily\footnotesize,morekeywords={-},morekeywords={+},keepspaces=true
]
 contract TicketOwnership is IERC721 {
   mapping(uint=>address) ticketToOwner;
   mapping(uint=>address) ticketApprovals;
   mapping(address=>mapping(address=>bool)) _operatorApprovals;
   
   function transferFrom(address _from, address _to, uint256 _tokenId) public {
     require(ticketToOwner[_tokenId] == ticketApprovals[_tokenId] && ticketToOwner[_tokenId] != msg.sender); 
     require(ticketToOwner[_tokenId] == _from);
     require(ticketToOwner[_tokenId] != address(0));
     require(_to != address(0));
+    require(_operatorApprovals[_from][msg.sender]); 
     _transfer(_from, _to, _tokenId);
   }
   function _transfer(address _from, address _to, uint256 _tokenId) internal {
     ticketToOwner[_tokenId] = _to;
     emit Transfer(_from, _to, _tokenId);
   }
   function approve(address _approved, uint256 _tokenId) public {
     require(msg.sender == ticketToOwner[_tokenId]);
     ticketApprovals[_tokenId] = _approved;
     emit Approval(msg.sender, _approved, _tokenId);
   }
   function isApprovedForAll(address owner, address operator) public view returns (bool) {
      return _operatorApprovals[owner][operator];
   }
   function ownerOf(uint256 _tokenId) external view returns (address) {
       require(ticketToOwner[_tokenId] != address(0));
       return ticketToOwner[_tokenId];
   }
   function getApproved(uint256 tokenId) external view returns (address) {
       require(ticketToOwner[tokenId] != address(0));
       return ticketApprovals[tokenId];
   }}
\end{lstlisting}
\vspace{-0.05in}
\mycaption{fig:721-high}{An ERC721 contract with a high-security
impact ERC violation.}
{(Code simplified for illustration.)
}
\end{center}
\end{minipage}
\vspace{-0.251in}
\end{figure}
}

\boldunderparagraph{ERC violations.}
Violating ERC rules could result in interoperability issues where the contracts may not work properly with wallets or DApps. 
It may also introduce security vulnerabilities, financial losses, or even token de-listing 
from exchanges that require compliance with ERC standards~\cite{erc-standard}.

Figure~\ref{fig:721-high} shows a real-world smart 
contract that violates a rule required by ERC721. The 
contract records NFT ownership (in this case, tickets) via \texttt{ticketToOwner} in line 2 and enables 
transfers through \texttt{transferFrom()} in lines 6--13. Under ERC721, an NFT may be transferred by its owner, an address approved for that specific NFT (\texttt{ticketApprovals} in line 3), or an address authorized to manage all of the owner’s 
NFTs (\texttt{\_operatorApprovals} in line 4). 
Consequently, \texttt{transferFrom()} must verify that \texttt{msg.sender} belongs to one of these categories.
In this implementation, the developers intentionally prevent owners from transferring tokens, allowing only a third-party platform to do so in line 7.
They also use the condition  ``\texttt{ticketToOwner[\_tokenId] == ticketApprovals[\_tokenId]}’’  in line 7 as a proxy for the owner’s intent to sell or transfer a ticket, after the owner calls \texttt{approve()} in lines 18--22 with his own address
as the first parameter. 
However, they omit a check for whether the caller is authorized to manage all of the owner’s tokens, enabling anyone to steal sellable tickets by transferring them to his own address.
The patch in line 11 fixes this issue by using \texttt{\_operatorApprovals} to verify whether the caller has the permission to manage all of the owner’s tickets. The \texttt{require} throws an exception and blocks the transaction if the check fails, 
thereby preventing unauthorized transfers 
and restoring ERC721 compliance.

\boldunderparagraph{State of the art.}
Following ERC rules is crucial, but developers often struggle due to the complexity of understanding all ERC requirements and 
corresponding contract code. 
ERC standards encompass numerous rules,
with 132 rules across the three ERCs we study.  
A single operation involves multiple rules. 
For example, ERC721 requires the 
public function \texttt{transferFrom()} of all ERC721 contracts 
complies with six rules, covering the API, 
caller privilege inspection, input validation, and logging. 
Failing to check caller privileges 
leads to the issue above. 
Moreover, ERC rules are often described in various formats, 
including both code declarations and natural language descriptions, 
which further complicates compliance.
Contract implementations are often complex, 
spanning hundreds to thousands of lines of source code across multiple files. 
Some details are obscured by intricate caller-callee relationships, 
while others involve numerous objects and functions, potentially written by different developers. 
Additionally, custom business logic (\eg, preventing NFT owners from transferring their own NFTs in Figure~\ref{fig:721-high}) further 
complicates the code. 
The complexities in both ERC rules and smart contracts make it difficult 
to ensure ERC compliance, leading 
ERC rule violations widely exist in real-world smart contracts \cite{humanaudited}.


Today's practices to avoid ERC violations are on three fronts. 
First, existing program-analysis tools can automatically verify certain smart contract criteria~\cite{slither-erc,erc20-verifier,ZepScope,VerX,ERCx,yang2023definition,liu2022finding, chen2019tokenscope, liu2022invcon,li2020securing}. 
However, their scope is limited, and they often struggle with complex ERC requirements. 
For example, none can detect the rule violation in 
Figure~\ref{fig:721-high} without expert-crafted validation 
rules. This limitation arises because many ERC rules involve semantic information (\eg, caller privileges) that is difficult to infer 
automatically, and some require contract-specific customization 
(\eg, locating where caller privileges are stored), 
a process often time-consuming or even infeasible.
Second, security 
experts  offer  auditing 
services~\cite{certik,revoluzion,pixelplex,blockhunters, immunebytes, antier,humanaudited}, providing more thorough 
validation than program-analysis tools.
However, these services are costly and time-consuming, 
requiring significant manual effort.
Third, some approaches rely solely on large language models (LLMs) to 
verify ERC compliance~\cite{GPT-RyanHarvey,GPT-MichaelDNorman}. 
However, these methods are prone to LLMs' hallucinations, often resulting in many false positives and false negatives.



\boldunderparagraph{Our proposal.}
This research aims to 
develop an effective, automated, and cost-effective
approach for verifying ERC rules. 
We begin with an empirical study of three widely used ERC standards, 
analyzing their 132 rules, 
focusing on the rules' nature, security risks of non-compliance, 
and their natural language descriptions.
Our study reveals four key insights valuable for Solidity developers, 
security analysts, and ERC protocol designers. 
Notably, about one-sixth of ERC rules verify whether an operator, token owner, or recipient has appropriate authority to perform specific operations.
Violations of such rules can open attack vectors and result in severe financial losses (\eg, Figure~\ref{fig:721-high}). We also find that most ERC rules
can be validated within a limited code scope (\eg, a function). Furthermore, we observe that although ERC rules cover diverse contract semantics, they are expressed via a limited set of linguistic patterns by ERC documents.

Building on insights from our empirical study, we develop \Tool{}, a fully automated tool for verifying smart contracts’ compliance with ERC standards. 
\Tool{} integrates the natural language understanding of LLMs with the formal guarantees of symbolic execution: it leverages an LLM to extract compliance rules from ERC documents, and then applies symbolic execution to check contracts against these rules. While the pipeline is straightforward in concept, \Tool{} tackles two key challenges in novel ways.
\textbf{\textit{Challenge I}}: Symbolic execution leverages input constraints
to determine when an ERC rule is violated. However, the space of possible constraints is enormous.
How can we leverage LLMs to generate the constraints representing ERC standard violations with minimal hallucinations while preserving full automation?
\textbf{\textit{Challenge II}}: Contracts may implement the same ERC differently, or even omit variables that encode essential ERC properties. How can we adapt symbolic execution to accommodate such variations effectively?

To address \emph{Challenge I}, we design a domain-specific language (DSL) 
that formalizes ERC rules using code constructs (\eg, contract fields) 
and execution events (\eg, token burning). 
Rather than asking the LLM to generate symbolic execution constraints directly, we first have it translate natural language rules into the DSL for each ERC.
We then apply deterministic program analysis to convert the translated rules into constraints for each contract.
This two-step process ensures the generation of constraints in a correct and consistent format.
Moreover, it restricts the LLM to a structured, narrow output space, thereby reducing 
variability and uncertainty in its responses. To our knowledge, 
we are the first to leverage an intermediate representation (\ie, the DSL)
to enhance LLMs’ performance when generating formal specifications.

%
To tackle \textit{Challenge II}, we introduce a set of constraint variables 
that capture the key semantic information required by ERC rules. 
For those that correspond to concrete contract program variables (\eg, 
the contract field indicating whether an address allows another to manage its tokens in Figure~\ref{fig:721-high}), we develop program analysis routines to automatically identify these contract variables across diverse contract implementations. 
For variables that represent abstract contract execution states 
without direct contract program counterparts (\eg, whether a function burns tokens), we define rules specifying how their values are updated 
during symbolic execution.
These designs significantly reduce discrepancies among different contract implementations and eliminate the need for contract-specific customization. 
By integrating these mechanisms, \Tool{} distinguishes itself from existing symbolic execution techniques~\cite{klee, S2E, SymCC}.
As far as we know, no prior work validate ERC requirements as many as ours.

We evaluate \Tool{} on three datasets: a large dataset of 4,000 contracts randomly selected from etherscan.io \cite{etherscan} and polygonscan.com \cite{polygonscan}, a ground-truth dataset of 40 contracts with violation labels (created by us), and a dataset of 20 contracts of two previously unstudied ERCs.
On the large dataset, \Tool{} detects 5,783 ERC rule violations, with 1,375 having a clear attack path leading to financial losses (one shown in Figure~\ref{fig:721-high}). 
The false discovery rate is 2.0\%. \Tool{} is effective in identifying ERC violations.
We compare \Tool{} against five static analysis 
techniques~\cite{slither,ZepScope,yang2023definition,ghaleb2023achecker,mythril}, 
a dynamic testing technique~\cite{ERCx}, 
and a human auditing service~\cite{humanaudited}, using the ground-truth dataset. 
\Tool{} detects over two times as many violations as all baselines 
while producing far fewer false positives, highlighting its superiority. 
Moreover, \Tool{} reduces both time and expenses 
by a factor of a thousand compared to the human auditing service, emphasizing its cost efficiency.
Finally, \Tool{} identifies all violations in the third dataset, 
showing strong generalization to ERCs beyond those we have studied.

In sum, we make the following contributions.


\begin{itemize}[noitemsep, topsep=0pt, leftmargin=.25in]
    \item We conduct the first empirical study on ERC rules made for smart-contract implementations.

    \item We propose a novel method using a DSL to integrate LLMs with symbolic execution, enabling the generation of properly formatted verification constraints. 

    \item We enhance ERC-rule verification by defining contract state variables and outlining procedures for updating them, addressing gaps in existing program variables.
    
    \item We design and implement \Tool to pinpoint ERC violations, and confirm its effectiveness, advancement, and generality via thorough experiments.

\end{itemize}


\section{Background}
This section gives the background of our project, 
covering Solidity smart contracts, ERCs, techniques related to ours, 
and the threat model under consideration.

\newcommand{\hlorange}[1]{\textcolor{myorange}{#1}}
\definecolor{myorange}{RGB}{255,140,0}

{
\begin{figure}[t]

\begin{minipage}{\columnwidth}
\begin{center}
\scriptsize
\begin{lstlisting}[numbers=left,xleftmargin=.20in, framexleftmargin=.16in,framexrightmargin=-0.05in,language=Solidity,basicstyle=\ttfamily,morekeywords={-},morekeywords={+},keepspaces=true,escapeinside={(*@}{@*)}]
/// @notice Transfer ownership of an NFT -- THE CALLER IS RESPONSIBLE
///  TO CONFIRM THAT `_to` IS CAPABLE OF RECEIVING NFTS OR ELSE
///  THEY MAY BE PERMANENTLY LOST
/// (*@\hlline{0.76}@*) @dev Throws unless `msg.sender` is the current owner, an authorized
/// (*@\hlline{0.52}@*) operator, or the approved address for this NFT. Throws if `_from` is
///  not the current owner. Throws if `_to` is the zero address. Throws if
///  `_tokenId` is not a valid NFT.
/// @param _from The current owner of the NFT
/// @param _to The new owner
/// @param _tokenId The NFT to transfer
function transferFrom(address _from, address _to, uint256 _tokenId) external payable;
\end{lstlisting}

\vspace{-0.1in}
\mycaption{fig:721-rule}{Natural language rules for ERC721's \texttt{transferFrom()} function.}
{(The rule violated in Figure~\ref{fig:721-high} is highlighted.)}
\end{center}
\end{minipage}
\vspace{-0.1in}
\end{figure}
}

\subsection{Ethereum and Solidity Smart Contracts}

Ethereum is a blockchain platform that allows developers to
write smart contracts for decentralized applications~\cite{eth-1,eth-2}. 
Users and
smart contracts are represented by distinct addresses to send and
receive Ether (the native cryptocurrency) and perform complex
transactions. Ethereum supports a vibrant digital economy, with
Ether prices exceeding \$3K and a total market value over \$300B~\cite{eth-price}.
Daily transactions surpass one million on Ethereum, representing \$4 billion in value~\cite{eth-daily}. 
Smart contracts play a central role in this ecosystem, as they govern most Ethereum’s transactions and provide the foundation for advanced functionalities~\cite{erc20,erc721,eth-defi}.

Solidity is the most widely used language for smart contracts~\cite{solidity-popular-1,solidity-popular-2}. 
Its syntax is similar to ECMAScript~\cite{ecmascript}, 
simplifying interactions with
the Ethereum system. Writing a contract in Solidity is similar to
defining a class in Java, featuring contract variables to store data
and functions to implement logic. Functions can be public, internal,
or private; public functions serve as the contract’s interface for external access and can be called by any user or contract, while private or internal functions do not.
Contracts can also define events, which are emitted 
during execution and are recorded on-chain for off-chain analysis. 

Figure~\ref{fig:721-high} illustrates an example contract. 
It contains three fields (lines 2–4) and two events 
defined in the \texttt{IERC721} interface, which are emitted in lines 16 and 21. 
Public function \texttt{transferFrom()} (lines 6--13)
can be invoked by 
any Ethereum user or contract, whereas the internal function \texttt{\_transfer()} (lines 14--17) is restricted to calls within the same contract.

\subsection{Ethereum Request for Comment (ERC)}
\label{sec:erc}

ERCs are technical documents that define how smart contracts 
should be implemented, ensuring they work consistently across different contracts, applications, and platforms, thus fostering the Ethereum ecosystem~\cite{erc-eip1, erc-standard, stefanovic2023proposal}. 

An ERC usually starts with a short motivation. For example, ERC721 introduces a standard interface that allows wallets and brokers to interact with NFTs on Ethereum~\cite{erc721}. It then provides a detailed specification, listing the required public functions and events along with their parameters, return values, and optional attributes. In addition, an ERC specifies requirements for each function or event through plain text or code comments before its declaration. 
For instance, Figure~\ref{fig:721-rule} shows a snippet of ERC721. 
It specifies that all ERC721-compliant contracts must implement the public function 
\texttt{transferFrom(address \_from, address \_to, uint256 \_tokenId)} to transfer the NFT identified by \texttt{\_tokenId} from \texttt{\_from} to \texttt{\_to}, along with four 
additional requirements on the function’s behavior: verifying the caller is authorized 
(either the owner, an address approved for the specific NFT, or an operator approved for all of the owner’s NFTs), checking \texttt{\_from} is the current owner, ensuring 
\texttt{\_to} is not the zero address, and confirming \texttt{\_tokenId} refers to a valid token.

Violating ERC rules can lead to serious financial losses and unexpected contract behavior. 
For instance, ERC721 requires \texttt{safeTransferFrom()} to call \texttt{onERC721Received()} when sending NFTs to a contract, 
and to check the return is a specific magic value. 
This ensures the receiver is able to handle the NFTs. Without this check, NFTs sent to an incompatible contract are permanently locked. 
Similarly, failing to verify that the caller is authorized to manage all of an owner’s tokens 
in
Figure~\ref{fig:721-high} enables attackers to steal any tradable tokens. In short, strict compliance with ERC rules is essential to safeguard assets and guarantee correct contract behavior.



\subsection{Related work}
\label{sec:related}


Several tools exist for detecting ERC rule violations, but their coverage is limited.
Slither~\cite{slither} provides dedicated checkers (\eg, arbitrary-send-erc20, slither-check-erc) to assess ERC compliance. 
These mainly verify the presence of required functions and events, check event emissions, and analyze contracts interacting with ERC-compliant contracts. 
However, they cannot capture complex rules, such as the one violated in Figure~\ref{fig:721-high}. 
The ERC20 verifier~\cite{erc20-verifier} performs checks similar to Slither but focuses only on ERC20.
AChecker~\cite{ghaleb2023achecker} identifies instances where access-control guards are entirely absent or where contract fields used in such guards can be freely modified by contract users.
ZepScope~\cite{ZepScope} derives rules from OpenZeppelin’s code and validates their use in contracts built on OpenZeppelin. Zepcompare~\cite{liu2025demy} detects vulnerabilities resulting from the use of buggy OpenZeppelin code.
VerX~\cite{VerX} checks whether smart contracts satisfy project-specific properties. While they address some ERC rules (\eg, access control), they miss many others (\eg, required event emissions).
NFTGuard~\cite{yang2023definition} detects only five error types in ERC721, 
far fewer than the full ERC specification.
Techniques for detecting inconsistencies or invariants also fall short: 
they either group message callers too coarsely~\cite{liu2022finding}, 
cover a subset of ERC-required functions~\cite{chen2019tokenscope, ghaleb2023achecker,fekih2023ercnft,ji2020deposafe}, 
assume partial correctness of code~\cite{liu2022invcon, beillahi2020behavioral}, 
or depend on expert-crafted rules~\cite{li2020securing,bram2021rich, nelaturu2023correctbydesign,keilty2022verimove,hajdu2020solcverify, abraham2019runtimeverification, chen2022declarative, shyamasundar2022framework,annenkov2021coqextract,certora}.
Some of the techniques (\eg, Certora~\cite{certora}, solc-verify~\cite{hajdu2020solcverify}, VERISOLID~\cite{nelaturu2023correctbydesign}) 
formalize only a limited subset of ERC rules, far fewer than those defined in the standards. To our knowledge, no prior work covers as many rules across ERC20, ERC721, and ERC1155 as we do.
ERCx~\cite{ERCx} checks compliance via unit 
tests, but the tests are incomplete and often fail to expose violations. Some customized 
ChatGPT-based auditing services~\cite{GPT-RyanHarvey,GPT-MichaelDNorman} have also emerged, but due to the large context of ERC 
documents, the size of contract code, and LLM hallucinations, they perform poorly when detecting 
ERC rule violations.

Researchers have also developed automated tools to 
identify other types of Solidity smart contract bugs, including reentrancy bugs~\cite{liu2018reguard, qian2020towards, xue2020cross, Oyente, Zeus,rodler2018sereum,tsankov2018securify,brent2018vandal, 10.1145/3238147.3238177,li2020securing, chen2024demystifying, zheng2023turn}, nondeterministic payment bugs~\cite{wang2019detecting, li20safepay}, consensus bugs~\cite{yang2021finding, chen2023tyr}, eclipse attacks~\cite{wust2016ethereum, xu2020eclipsed, marcus2018low},
out-of-gas attacks~\cite{grech2018madmax,ghaleb2022etainter}, errors in DApps~\cite{VetSC},
cryptographic errors~\cite{zhang2024demystifying}, 
accounting errors~\cite{zhang2024towards}, state-reverting errors~\cite{10.1145/3597926.3598111,10.1145/3395363.3404366},
documentation errors~\cite{zhu2022identifying}, centralization errors~\cite{lin2024definition, ma2023piedpiper}, unprotected self-destruction~\cite{krupp2018teether, mythril, ye2022vulpedia, chang2019scompile}, integer bugs~\cite{torres2018osiris,tikhomirov2018smartcheck,nguyen2020sfuzz,chen_numscout_2025,10.1145/3656416,8952204,yu2025smart,ayoade2019bytecoderewriting}, oracle manipulation attacks~\cite{deng2024safeguarding, qin2025enhancing}, flash loan attacks~\cite{chen2024flashsyn}, event-ordering bugs~\cite{kolluri2019laws}, fairness issues~\cite{liu2020fairness}, and exploitable errors~\cite{zhang2023demystifying}.
However, these techniques do not focus on ERC-specific 
semantics and cannot detect ERC rule violations. 

Security experts audit smart contracts to detect 
vulnerabilities and 
logic flaws~\cite{certik,revoluzion,pixelplex,blockhunters,immunebytes,antier,humanaudited}, with some also verifying ERC compliance. However, these services 
are costly and time-consuming, making them less appealing to 
developers than automated tools.

Prior work has applied LLMs (or machine learning) to analyze Solidity code for 
vulnerability detection~\cite{sun2023gpt, ma2024combiningfinetuningllmbasedagents, propertygpt, sendner2023smarter, hao2023smartcoco, lin2025promfuzz, zhong2025detecting, luo2024scvhunter}, bug fixing~\cite{ibbaleveraging, yuan2025leveraging}, bug reproduction~\cite{wei2025veriexploit}, and
exploit generation~\cite{wu2024advscanner,sun2025learning}.
Others have leveraged LLMs to generate verification specifications for Rust programs~\cite{autoverus,chen2024automated}. 
In contrast, our goal is to validate contracts against a broad range of ERC implementation rules, tackling a problem distinct from these prior techniques.


In summary, existing program-analysis-based 
techniques either provide limited ERC compliance checks 
or target unrelated bugs. 
Manual audits are thorough but are costly and 
time-consuming. LLM-based approaches show promise for 
general error detection but struggle to identify 
specific ERC rule violations. Our research overcomes 
these limitations by developing a fully automated, end-to-end solution for verifying compliance with a substantial fraction of ERC rules.


\subsection{Threat Model}
In our study, attackers are not required to obtain elevated privileges (\eg, compromising the Ethereum network, stealing private keys). 
Instead, as long as they have sufficient gas to invoke public functions of smart contracts deployed on Ethereum, 
they are capable of exploiting ERC rule violations to launch attacks.



\section{Empirical Study on ERC Rules}
\label{sec:study}



From the 102 finalized ERCs, 
we select ERC20~\cite{erc20}, ERC721~\cite{erc721}, and ERC1155~\cite{erc1155} 
as our study targets. 
They are technical standards for fungible tokens (\eg, cryptocurrencies),
non-fungible tokens (NFTs), and contracts managing both.
Our selection is based on their popularity,  
their complexity, 
and their significance in the Ethereum ecosystem~\cite{ erc20-popular,USDT,SHIB,Binance, opensea,rarible,Horizon,9Lives,Reewardio}.
For example, in the last 180 days, around 3.6 million contracts implementing ERCs were deployed, including 2.6 million for ERC20, 
0.6 million for ERC721, and 0.3 million for ERC1155.

We carefully review the specification section of each ERC official document 
and manually identify a natural language sentence as a rule if it is related to contract implementations, 
specifies a clear restricting target, and provides actionable checking criteria. 
Certain rules explicitly use terms like ``must'' or ``should'' to convey obligations.
%
We identify a total of 132 rules across the three ERCs: 32 
from ERC20, 60 from ERC721, and 40 from ERC1155. 
%

Our study primarily answers three
key questions regarding the identified rules: 1) what rules are
specified? 2) why are they specified? and 3) how are they
specified in natural language? The goal is to
garner insights for building techniques that can automatically detect
rule violations. 
To ensure objectivity, two authors independently analyze each ERC and 
then discuss their findings to resolve any disagreements.
In sum, the empirical study takes roughly four human-weeks of effort. 

\begin{table}[t]
\centering
\footnotesize

\mycaption{tab:study}
{ERC rules' content and security impacts.}
{
}
{

\begin{tabular}{|l|c|c|c||c|}
\hline
\diagbox{\textbf{content}}{\textbf{impact}}
                 &  {\textbf{High}} & {\textbf{Medium}} & {\textbf{Low}}  & {\textbf{Total}} \\

\hline
\hline

{\textbf{Privilege Check}}     & 24  &  0 &   0  &  24   \\ \hline
{\textbf{Functionality}}       & 12  & 30  &   0  & 42  \\ \hline
{\textbf{API}}                 & 0   &  33 &   0  & 33 \\ \hline 
{\textbf{Logging}}             & 0   &  0 &  33 & 33 \\ \hline 
\hline
{\textbf{Total}}               & 36 &  63 &  33 & 132 \\ \hline
\end{tabular}
}
\vspace{-0.1in}
\end{table}

\subsection{Rule Content (What)}
\label{sec:what}
An ERC rule generally requires a public function to include a specific piece of code. 
Based on the semantic nature of the code, we categorize the rules into four groups, as shown in Table~\ref{tab:study}.

\indentitalicparagraph{Privilege Checks.} 
Regarding the required code patterns, 
20 rules involve 
checking a condition and throwing an exception if it fails, 
while others require calling a function for a subsequent check.
For the object being checked in each rule,
10 rules pertain to the operator (\eg, \texttt{msg.sender}) of a token operation.  
For example, the implementation in Figure~\ref{fig:721-high}
violates the highlighted rule in Figure~\ref{fig:721-rule},
which requires verifying \texttt{msg.sender} is authorized to
transfer the NFT.
Additionally, three rules address whether the token owner holds sufficient privileges.
For example, 
ERC1155 mandates \texttt{safeTransferFrom()} only executes when 
the source address holds enough tokens.
The remaining 11 rules focus on transfer recipients. 
For instance, ERC721 prohibits \texttt{transferFrom()} from sending 
NFTs to the zero address (line 6 in Figure~\ref{fig:721-rule}). 
It also requires calling \texttt{onERC721Received()} 
if the recipient is a contract, checking whether the return value 
matches a magic number, and throwing an exception if it does not.

\indentitalicparagraph{Functionality Requirements.} 
Five types of code are required by rules in this category.
First, 24 rules specify how functions should generate return values. 
For example, ERC1155 requires public function 
\texttt{balanceOf(address \_owner, uint256 \_id)} to return the amount of tokens of type \texttt{\_id} 
owned by \texttt{\_owner}. 
In particular, when a
function returns a Boolean value, 
it implicitly requires returning \texttt{true} on success
and \texttt{false} otherwise.
Second, 12 rules address how to validate input parameters and when to throw exceptions. 
For instance, ERC20 dictates that \texttt{transferFrom()} treats 
zero-token transfers the same as non-zero transfers. 
Third, two rules explicitly mandate the associated function to 
throw an exception when any error occurs. 
Fourth, three rules specify how to update particular variables. 
For instance, one ERC20 rule requires \texttt{approve(address \_spender, uint256 \_value)} to overwrite the 
allowance value that the message caller allows \texttt{\_spender} to
manipulate with \texttt{\_value}. The remaining rule is from 
ERC1155 which allows transferring multiple token types in one transaction  
but requires the balance update for each input token type to 
follow their order in the input array.

\indentitalicparagraph{API Requirements.}
The three ERCs mandate 33 public functions for 
contract interaction. To ensure compatibility, 
developers must implement these APIs as specified.


\indentitalicparagraph{Logging.}
ERCs enforce logging by requiring event emissions. 
For example, ERC721 mandates emitting a \texttt{Transfer} event whenever a transfer occurs 
(\eg, line 16 in Figure~\ref{fig:721-high}). In total, 33 rules govern logging: 
24 specify when to emit events, eight further define required parameters, 
and nine address event declarations.

\stepcounter{insight}
\indentboldunderparagraph{Insight \arabic{insight}:}
{\it{
ERC rules encompass diverse contract semantics, making it challenging to develop program analysis techniques that cover these semantics and detect rule violations.
}}

We further study the valid scope for each rule. 
Among the 132 rules, 
106 rules are confined to a single function. 
For instance, ERC721 mandates \texttt{transferFrom()} 
in Figure~\ref{fig:721-high} scrutinizes whether the message caller 
is authorized to handle the token owner's tokens 
(the highlighted line in Figure~\ref{fig:721-rule}). 
Moreover, nine rules pertain to event declarations. 
The valid scopes of the remaining 17 cases encompass the entire contract.
For instance, 
for every token transfer, both ERC20 and ERC721 mandate emitting a \texttt{Transfer} event. 


\stepcounter{insight}
\indentboldunderparagraph{Insight \arabic{insight}:}
{\it{
Most ERC rules can be checked within a function or at a declaration site, and there is no need to analyze the entire contract for compliance with these rules. 
}}

Insight 1 focuses on the requirements mandated by ERC rules, 
while Insight 2 highlights the code regions where these requirements should be inspected. 
They represent two orthogonal dimensions of ERC rule validation. Although many ERC rules can be checked within a limited scope, 
their validation remains challenging due to the wide variety of contract semantics involved.

\begin{table}[t]
    \centering
    \footnotesize
    \setlength{\tabcolsep}{3pt} 
    \mycaption{tab:linguistic}
    {Linguistic Patterns.}
    {([*]: an optional parameter. 
    Subscript [root] marks the root of a sentence.
    Table 7 in the appendix lists all possible words for each symbol.
    )
    }
    \begin{tabular}
    {|l|l|c|}
    \hline
 \textbf{ID}   & \textbf{Patterns} &  \textbf{Total} \\ \hline \hline
    TP1 & [SUB] [MUST] THROW\textsubscript{[root]} COND       & 28        \\ \hline
    TP2 & ACTION MUST THROW\textsubscript{[root]}                          & 2         \\ \hline
    TP3 & CALLER MUST APPROVE\textsubscript{[root]} ACTION           & 2       \\ \hline
    TP4 & ACTION BE\textsubscript{[root]} INVALID                  & 2      \\ \hline

    CP1 & COND SUB MUST CALL\textsubscript{[root]} SUB [, VAR MUST ASSIGN VALUE]  & 4    \\ \hline
    EP1 & [EVENT] [MUST] EMIT\textsubscript{[root]} [COND] [, VAR [MUST] ASSIGN [PREP] VALUE]                       &     15    \\ \hline

    EP2 & [ACTION] [MUST] EMIT\textsubscript{[root]} EVENT [PREP VAR ASSIGN VALUE COND]                       & 9         \\ \hline
    RP1 & SUB [MUST] RETURN\textsubscript{[root]} VALUE [COND]  &   24 \\ \hline
    AP1 & [SUB] [MUST] ASSIGN\textsubscript{[root]} VALUE [PREP] VALUE           &  2        \\ \hline
    AP2 & VALUE [MUST] ASSIGN\textsubscript{[root]} PREP VALUE           &  1        \\ \hline
    OP1 & ACTION MUST FOLLOW\textsubscript{[root]} ORDER                 &  1        \\ \hline \hline

{\textbf{Total}}  &  &  90 \\ \hline

\end{tabular}
\vspace{-0.20in}
\end{table}

\subsection{Violation Impact (Why)}

We analyze the security risks of rule violations to understand the rules’ rationale, 
and categorize their 
impacts into three levels, as shown in Table~\ref{tab:study}.

\indentitalicparagraph{High.}
A rule is considered high-impact if violating it allows attackers to craft malicious 
inputs for a public function to trigger unauthorized token transfers, 
incorrect token balances or allowances, or permanent token loss. Such violations present a direct attack path leading to financial losses. 
As shown in Table~\ref{tab:study}, 36 rules fall into this category, 
including all rules related to privilege checks. 
For example, failing to verify an operator’s permissions allows NFT theft (\eg, Figure~\ref{fig:721-high}), while not checking a recipient address is non-zero causes tokens to be lost forever.

Among the functionality requirement rules, violating 12 can also lead to financial losses.
Of these, nine rules outline how to generate return values 
representing token ownership or privileges to operate tokens, 
such as \texttt{balanceOf()} returning 
an address’s token balance.
Errors that provide incorrect returns for 
these functions can trap tokens in an address
or allow unauthorized token manipulation. 
The remaining three rules ensure proper token ownership updates. 
For instance, ERC20 requires function \texttt{approve(address \_spender, uint256 \_value)} 
to overwrite the amount of tokens \texttt{\_spender} authorizes the message caller
to manipulate with \texttt{\_value}.


%
%
%

\indentitalicparagraph{Medium.}
A rule has a medium impact if its violation causes 
unexpected contract or transaction 
behavior but does not create a direct path to 
financial losses. 
For example, if a public function's API fails to meet its ERC declaration, 
invoking the function with a message call following the requirement would trigger an exception. Another example is ERC20's 
requirement that \texttt{transferFrom()} treats zero-token 
transfers the same as non-zero transfers. If this rule is violated, the contract's behavior would 
be unexpected for the message caller.

\indentitalicparagraph{Low.}
All event-related rules are about logging. We consider their security impact as low.

\stepcounter{insight}
\indentboldunderparagraph{Insight \arabic{insight}:}
{\it{
For numerous rules, their violations present a clear attack path 
for potential financial losses, emphasizing the urgency of detecting and 
addressing these violations.
}}

\subsection{Linguistic Patterns (How)}
\label{sec:patterns}

Of the 132 rules, 42 specifically address function or event declarations 
using Solidity source code. 
The remaining 90 are described 
in natural language. 
As shown in Table~\ref{tab:linguistic}, we identify 11 linguistic patterns 
used to express the 90 rules.
These patterns correspond to six types of code implementations, 
as indicated by their ID prefixes (column ID in Table~\ref{tab:linguistic}): 
TP indicates throwing or not throwing an exception under certain conditions; 
CP involves calling a function, possibly with specific argument requirements; 
EP denotes emitting an event under certain conditions, 
potentially with argument requirements; 
RP involves returning a required value under certain conditions; 
AP refers to updating a variable with a new value; 
and OP is for performing actions in a specific order.

Out of the 11 patterns, 
four (TP1, EP1, EP2, and RP1) account for over 80\% of the rules. 
TP1 is primarily used for privilege checks. For example, the rule highlighted in Figure~\ref{fig:721-rule}
falls under TP1.
EP1 and EP2 are used to emit events, 
while RP1 specifies return values. In contrast, the remaining patterns 
cover many fewer rules.
For instance, AP2 and OP1 are each associated with only one rule.

\stepcounter{insight}
\indentboldunderparagraph{Insight \arabic{insight}:}
{\it{
Most ERC rules follow common linguistic patterns, while a few are uniquely specified.
}}

\begin{figure}[t]
\centering
\includegraphics[width=0.7\columnwidth]{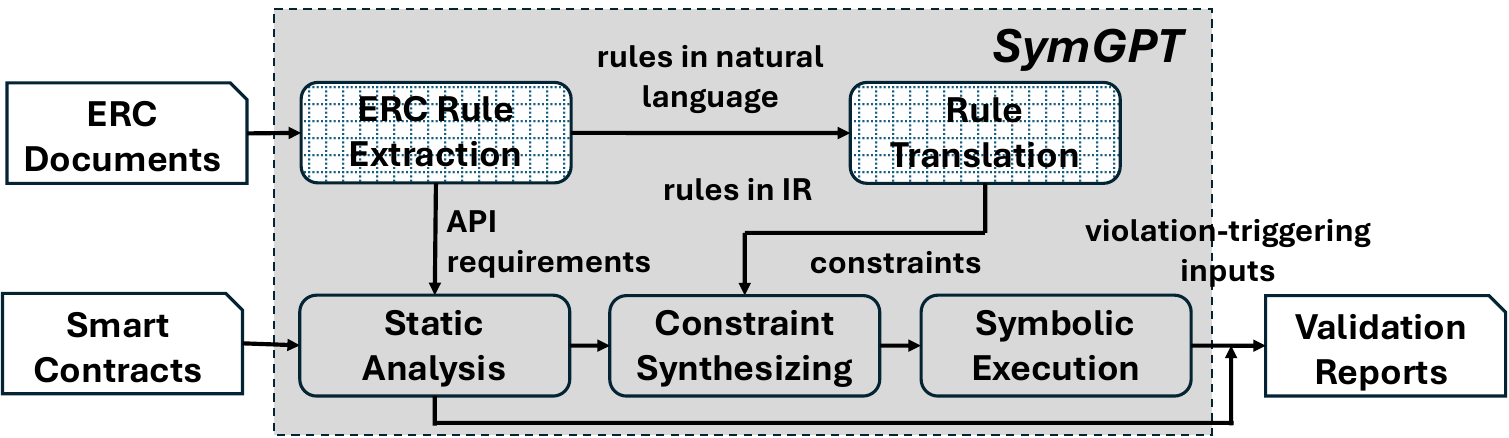}
\mycaption{fig:workflow}{The workflow of \Tool{}.}
{(Components with a pattern background are powered by an LLM. 
Example outputs for 
Rule Translation and Constraint
Synthesizing are shown in Figures 
\ref{fig:721-high-grammar} and \ref{fig:gen-3}.)
}

\end{figure}

\section{The Design of \Tool{}}
\label{sec:tool}

\Tool{} takes as input ERC documents written in natural language and smart contract source code.
It automatically determines whether and where 
the contracts violate any rules defined in the ERCs, \emph{without any human intervention.}
This makes it useful during in-house development for ensuring ERC compliance before deploying contracts on Ethereum.

Figure~\ref{fig:workflow} illustrates \Tool{}’s workflow and architecture,
which consists of five components. 
The two patterned components process
ERC documents by \emph{extracting} ERC rules from them 
(Section~\ref{sec:extraction}) and \emph{converting} 
those rules into an intermediate representation (IR) (Section~\ref{sec:translation}).
Since ERC rules encompass diverse aspects of contract semantics (see Insight 1 in Section~\ref{sec:study}), manually extracting and translating 
them is both tedious and error-prone.
We therefore employ an LLM  to automate these two stages, leveraging its strong natural language understanding capabilities~\cite{yang2024harnessing}.
This fully automated pipeline also simplifies extending \Tool{} to support new ERCs (see Section~\ref{sec:3525}). 
The remaining three components analyze each individual contract by 
1) checking that function and event declarations comply with ERC requirements, 
while gathering information for the next step (Section~\ref{sec:se});
2) translating ERC rules from the IR into contract-specific, rule-violation constraints (Section~\ref{sec:synthesizing}); 
and
3) running symbolic execution to 
determine whether the contract satisfies these constraints, 
and reporting any violations (Section~\ref{sec:se}).




\begin{figure}[t]
\centering
\begin{tcolorbox}[
  enhanced,
  colback=promptbg,
  colframe=promptborder,
  boxrule=0.3pt,
  arc=2mm,
  left=2mm,
  right=2mm,
  top=1mm,
  bottom=1mm,
  drop shadow,
  width=1\linewidth
]
\begin{lstlisting}[
  language=GPTPrompt,
  basicstyle=\ttfamily\footnotesize,
  numbers=none,
  frame=none,
  columns=fullflexible,
  keepspaces=true,
  showstringspaces=false
]
Given an ERC, which contains a list of functions, events, and rules. Each function has its own rules, usually above their Solidity declaration. Rules indicating throw or check usually have patterns like
"""
[sub] [must] throw condition
action must throw
caller must be approved action
action are considered invalid
"""
Other patterns that serve similar purposes also count.

Rules:"""
{{description}}
{{API}}
"""
List the conditions that need to be thrown or checked in the given rules for function {{API}} in a JSON array with the following format:

if: <string, condition to throw or check>
throw: <boolean, whether it needs to throw or should not be thrown>

If there is no such rule, simply left empty.
\end{lstlisting}
\end{tcolorbox}

\vspace{-0.4em}
\mycaption{fig:prompt1}{The template for prompts extracting TP rules.}{
(``\{\{description\}\}'': text descriptions precede a function declaration in an ERC document. ``\{\{API\}\}'': the declaration of the function.)
}
\vspace{-0.15in}
\end{figure}

The rest of this section follows the workflow to explain \Tool{}’s technical details in depth.

\subsection{ERC Rule Extraction}
\label{sec:extraction}

\Tool{} focuses on rules outlined in ERC documents.
While some correctness and performance 
rules can be identified and extracted from smart 
contract code~\cite{ZepScope,mengting-gas}, 
validating compliance with them is beyond this paper's scope.
Furthermore, 
the experimental results in Section~\ref{sec:compare} show that the 
rules extracted from contract code cover only a small subset of ERC requirements.

A naïve way to extract rules from an ERC document 
with an LLM 
is to provide the entire document and ask the LLM to identify all rules. 
However, this often leads to incomplete or inaccurate extraction. 
Instead, we break each ERC document into subsections and
process them separately. 
Specifically, since each ERC’s rules are detailed in the specification section and 
precede the relevant function or event declarations, 
we use regular expressions 
to isolate the specification section and split the section 
into smaller parts, ending at each function or event declaration. 
We then instruct the LLM to analyze 
each declaration along 
with its preceding text description.

We design prompts based on the linguistic patterns in Table~\ref{tab:linguistic}. 
Each prompt begins with an introduction, followed by a set of linguistic patterns sharing the same ID prefix. 
It then presents the text description 
and declaration of a function (or an event) and asks the LLM to extract rules 
from the description, including any relevant value requests 
(\eg, emitting an event with a specific parameter), based on the patterns. 
Finally, the prompt explains the JSON schema for formatting the extracted rules.
Including linguistic patterns helps the LLM better 
understand what ERC rules are, improving extraction accuracy.
%
%
%
Figure~\ref{fig:prompt1} illustrates the prompt 
template used for rules in the TP linguistic-pattern group. 
The rule violated in Figure~\ref{fig:721-high} can be accurately extracted using a 
prompt instantiated from this template, where the 
placeholders ``\{\{API\}\}'' and ``\{\{description\}\}'' are 
replaced with the declaration of \texttt{transferFrom()} and the textual description preceding it in the ERC document, both of which are shown in Figure~\ref{fig:721-rule}.
While summarizing linguistic patterns requires effort, 
most rules follow a limited number 
of patterns (see Insight 4 in Section~\ref{sec:patterns}), making
the patterns likely to 
apply to ERCs that we have not studied. 

{
\begin{figure}[t]

\begin{minipage}{\columnwidth}
\begin{center}
\scriptsize
\begin{lstlisting}[numbers=left, xleftmargin=.20in, framexleftmargin=.16in,framexrightmargin=-0.05in,language=ebnf,
basicstyle=\ttfamily\footnotesize,keepspaces=true]
<check> ::= <throw> | <call> | <emit> | <assign> | <follow>
//Lines 3--7: five top-level non-terminals
<throw> ::= if <b_exp> then checkThrow(<fun>, <flag>)
<call> ::= if <b_exp> then checkCall(<fun>, <fun>) [with <b_exp>]
<emit> ::= if <b_exp> then checkEmit(<fun>, <event>) [with <b_exp>]
<assign> ::= checkEndValue(<fun>, <value>, <value>)
<follow> ::= checkOrder(<op>, <op>, <op>, <op>)

<b_exp> ::= not <b_exp> | <b_exp> (and | or) <b_exp> | <value> <b_op> <value> | <change> | <mint> | <burn>
<value> ::= msg.sender | <field_value> | <para> | ...
<field_value> ::= getFieldValue(<field> [,<key>]*)
<field> ::= getField(<anchor_fun>)
<para> ::= getPara(<fun> | <event>, <index>)
<change> ::= checkChange(<fun>, <field_name> [, <key>]*)
<mint> ::= checkMint(<fun>, <field_name>)
<burn> ::= checkBurn(<fun>, <field_name>)




\end{lstlisting}
\vspace{-0.1in}
\mycaption{fig:ebnf}{The EBNF grammar of the domain-specific language.}
{(Terminals in blue are literal tokens and those in violet are utility functions. The grammar is simplified due to limited space.)}
\end{center}
\end{minipage}
\end{figure}
}

\subsection{ERC Rule Translation}
\label{sec:translation}
Generating constraints directly from natural-language rules poses significant challenges for LLMs, primarily for two reasons.
First, the vast space of possible constraints introduces substantial uncertainty and 
increases the likelihood of hallucinations.
Second, different contracts implement the same ERC in diverse ways, 
leading some rules to depend on contract-specific details to be properly validated. Generating constraints for such rules would either require costly 
per-contract analysis by the LLM or risk reduced accuracy if relevant information is omitted.

To overcome these challenges, we define a domain-specific language (DSL)
as an intermediate representation (IR) for ERC rules 
and prompt the LLM to translate each rule into this IR. 
We then perform program analysis on individual contracts to instantiate contract-specific constraints from the IR (Section~\ref{sec:synthesizing}), thereby removing the need for the LLM to analyze each contract individually.
We adopt extended Backus--Naur form (EBNF) to specify the grammar of the DSL,
because ERC rules primarily describe interface-level behaviors and control-flow requirements, both of which can be naturally expressed using EBNF. 
Other alternatives (\eg, regular expressions, Datalog) are either insufficiently expressive to capture the semantics involved in ERC rules 
or excessively powerful, introducing unnecessary complexity.

\bolditalicparagraph{Domain-Specific Language.}
Figure~\ref{fig:ebnf} shows the language grammar, which iteratively expands the non-terminals
(symbols enclosed by ``<>’’) into terminals (those not enclosed by ``<>’’). 
The grammar contains five top-level non-terminals in line 1,
corresponding to five linguistic-pattern 
groups in Table~\ref{tab:linguistic}. 
All groups are covered 
except the one related to return value generation.
This omission stems from the difficulty of using contract elements explicitly required by ERCs to define the required return values. For instance, 
ERC20 requires the \texttt{name()} function of every ERC20 contract to return the name of the token the contract manages, 
but does not mandate where the name must be stored, making it difficult to formalize this rule across diverse ERC20 implementations.

The terminals include contract elements 
(\eg, public functions and their parameters) defined by ERCs, 
utility functions (shown in violet in Figure~\ref{fig:ebnf}), 
and literal tokens (\eg, \texttt{if}, \texttt{and}, shown in blue in Figure~\ref{fig:ebnf}).
Utility functions are analysis routines defined by us. 
Their names are carefully chosen to reflect their exact functionalities, 
aiding the LLM in understanding them 
and performing the formalization. For example, \texttt{checkThrow(<fun>, <flag>)} 
returns a Boolean value, indicating whether contract function \texttt{<fun>} throws an exception or not as indicated by \texttt{<flag>}.

\indentitalicparagraph{Non-terminal <throw>.}
\texttt{<throw>} in line 3 formalizes 
how to check rules requiring a function to throw or not throw an exception under a specific condition. 
For one such rule, 
both non-terminal \texttt{<b\_exp>}, representing the condition, 
and the parameter of the utility function \texttt{checkThrow(<fun>, <flag>)} need to be specified.

How to extend \texttt{<b\_exp>} is shown in lines 9--16. 
In line 9, a \texttt{<b\_exp>} can be a compound Boolean expression, 
a comparison between two \texttt{<value>}s, 
or the returns of three utility functions. 
Line 10 denotes
a \texttt{<value>} can represent 
\texttt{msg.sender}, a contract field's value, or other similar elements. 
Utility function \texttt{getFieldValue()} in line 11 retrieves a contract field’s value, taking 
the field name and optional \texttt{key}s (for 
array or mapping) as input.
ERCs do not impose any requirements on field names 
but do require the names of certain public functions that return the values of contract fields. 
We consider these functions as anchor functions to identify field names. For example,
function \texttt{isApprovedForAll()} in lines 23--25 of Figure~\ref{fig:721-high} is explicitly required by ERC721, and it is the anchor function
to identify the contract field tracking 
whether one address allows another
to manage all its tokens. 
We treat all public, read-only 
functions required by ERCs as 
potential anchor functions, and request the LLM to determine 
any specific anchor function for each rule when 
it translates the rule into the DSL.
Utility function \texttt{getField()} in line 12 takes an anchor function as 
input and outputs the name of the field returned by the anchor function.


For example, Figure~\ref{fig:721-high-grammar} 
shows the rule violated in 
Figure~\ref{fig:721-high} in the DSL. This rule applies to 
contract function \texttt{transferFrom()}, whose 
declaration appears in line 1. In lines 2, 3, and 4, the LLM 
identifies contract functions 
\texttt{ownerOf()}, 
\texttt{getApproved()}, 
and 
\texttt{isApprovedForAll()} 
as the anchor functions 
for the contract fields 
\texttt{ticketToOwner}, \texttt{ticketApprovals}, and 
\texttt{\_operatorApprovals}, respectively. The utility 
function \texttt{getField()} analyzes these contract 
functions to extract the associated field names.
Since these fields are mappings, utility 
function \texttt{getFieldValue()} requires one or 
two additional keys as parameters. The condition checks whether \texttt{msg.sender} is the 
token owner, the address approved for the token, or the address authorized to manage all of the owner’s 
tokens. If \texttt{msg.sender} does not satisfy any of these cases, \texttt{transferFrom()} must throw 
an exception, which is verified by the utility function \texttt{checkThrow()} in line 6.
{
\begin{figure}[t]

\begin{minipage}{\columnwidth}
\footnotesize
\begin{lstlisting}[numbers=left,framexleftmargin=.16in,xleftmargin=.20in,framexrightmargin=-0.05in,language=ebnf,basicstyle=\ttfamily,morekeywords={-},morekeywords={+},keepspaces=true]
//function transferFrom(address _from, address _to, uint256 _tokenId)
if msg.sender != getFieldValue(getField(`ownerOf()`), getPara(`transferFrom()`, 2))
   and msg.sender != getFieldValue(getField(`getApproved()`), getPara('transferFrom()', 2)) 
   and not getFieldValue(getField(`isApprovedForAll()`), getPara(`transferFrom()`, 0), msg.sender)
then 
  checkThrow(`transferFrom()`, true);
\end{lstlisting}
\vspace{-0.1in}
\mycaption{fig:721-high-grammar}{The rule violated in Figure~\ref{fig:721-high} in the DSL.}
{}
\end{minipage}
\vspace{-0.2in}
\end{figure}
}




\begin{figure}[t]
\centering
\begin{tcolorbox}[
  enhanced,
  colback=promptbg,
  colframe=promptborder,
  boxrule=0.3pt,
  arc=2mm,
  left=2mm,
  right=2mm,
  top=1mm,
  bottom=1mm,
  drop shadow,
  width=1\linewidth
]
\begin{lstlisting}[
  language=GPTPrompt,
  basicstyle=\ttfamily\footnotesize,
  numbers=none,
  frame=none,
  columns=fullflexible,
  keepspaces=true,
  showstringspaces=false
]
For {{API}}
rule: {{rule}}
{{args}}
By using the following JSON schema of the configuration for the rule verification:
{{DSL}}
{{anchors}}
Generate the JSON for the rule. 
\end{lstlisting}
\end{tcolorbox}

\vspace{-0.4em}
\mycaption{fig:prompt2}{The template for prompts translating a rule in natural language into the DSL.}{(``\{\{API\}\}'': the source-code declaration of a function or an event. 
``\{\{rule\}\}'': an extracted natural language rule.
``\{\{args\}\}'': possible arguments for the rule.
``\{\{DSL\}\}'': the JSON schema of the top-level non-terminal.
``\{\{anchors\}\}'': the list of all public, read-only functions required by the ERC.
)
}
\end{figure}

\indentitalicparagraph{Non-terminals <call> \& <emit>.}
These two check whether a contract function 
calls another function or emits an event under specific conditions. 
They also ensure that the parameters of the called function 
or emitted event meet the required criteria. 
For example, ERC1155 requires a function to emit
a \texttt{TransferSingle} event with the event's second 
parameter set to zero when minting tokens. 
To enforce this rule, we use the utility function \texttt{checkMint(<fun>)} in line 15 of Figure~\ref{fig:ebnf}
as the first \texttt{<b\_exp>} of \texttt{<emit>} to verify that the analyzed 
contract function only increases token balances without decreasing them, 
indicating a token minting action. Once confirmed, we then check if the 
contract function emits the required event using \texttt{checkEmit(<fun>, <event>)}. 
If this is also confirmed, we proceed to verify that the second parameter 
of the emitted event is zero, specified by the second \texttt{<b\_exp>}.

\indentitalicparagraph{Non-terminal <assign>.}
\texttt{<assign>} leverages utility function 
\texttt{checkEndValue(<fun>, <value>, <value>)} 
in line 6 of Figure~\ref{fig:ebnf} to perform an unconditional check.
The utility function inspects whether the first \texttt{<value>} matches 
the second at the end of contract function \texttt{<fun>}.

\indentitalicparagraph{Non-terminal <follow>.}
\texttt{<follow>} checks whether two pairs of operations are in the same order
using utility function \texttt{checkOrder()}.

\bolditalicparagraph{Prompting the LLM.}
Some ERC rules apply directly to function and event declarations 
and can be inspected without translation.  
For other rules, we use the LLM to translate each one individually. 
Figure~\ref{fig:prompt2} shows the prompt template. 
It begins with a function or event's source code declaration, 
followed by an extracted natural language rule for the function or event 
and any possible arguments. 
The template then specifies the JSON schema for the top-level non-terminal of 
the rule to help the LLM understand the grammar of the DSL. 
Next, the template introduces all public, read-only functions required by the ERC as possible anchor functions.
Finally, it instructs the LLM to translate the rule into the DSL, 
according to the JSON schema.

\begin{table}[t]
\centering
\footnotesize
\mycaption{tab:step1}
{Translations from utility functions into constraints.}
{
}


{
\begin{tabular}{|p{0.38\linewidth}|p{0.50\linewidth}|}
\hline
 {\textbf{Utility Functions}} & {\textbf{Constraints}} \\

\hline
\hline

\texttt{checkThrow(<fun>, <flag>)} & $\var{TH} = \var{flag}$                   \\ \hline
\texttt{checkCall(<fun>, <fun2>)}  & ${\var{CA}_\var{fun2}} = \var{true}$                \\ \hline
\texttt{checkEmit(<fun>, <e>)}     &    ${\var{EM}_\var{e}} = \var{true}$               \\ \hline
\texttt{checkEndValue(<fun>,<v1>,<v2>)} & $v_1 = v_2$                 \\ \hline
\texttt{checkChange(<fun>, <f>}) &  ${\var{BC}_\var{f}}=\var{true}$                         \\ \hline
\texttt{checkMint(<fun>, <f>)} &    $({\var{BI}_\var{f}}\land \neg {\var{BD}_\var{f}}) =\var{true}$            \\ \hline
\texttt{checkBurn(<fun>, <f>)}  &  $(\neg {\var{BI}_\var{f}}\land {\var{BD}_\var{f}}) = \var{true}$ \\ \hline
\texttt{checkOrder(<op1>,<op2>,<op3>,<op4>)} & $(O_\var{op1} < O_\var{op2} \land O_\var{op3} < O_\var{op4}) \lor (O_\var{op1} > O_\var{op2} \land O_\var{op3} > O_\var{op4})$ \\ \hline

\texttt{getField(<anchor\_function>)} &  contract field returned by \texttt{<anchor\_function>}  \\ \hline
\texttt{getPara(<fun>, <i>)} & the ith formal parameter of \texttt{<fun>}\\ \hline 
\texttt{getFieldValue(<f>(, <key>)*)} & the value of \texttt{<f>} or the element of \texttt{<f>} indexed by \texttt{<key>} \\ \hline

\end{tabular}
}
\end{table}

\subsection{Synthesizing Violation Constraints}
\label{sec:synthesizing}

This component analyzes individual 
contracts and translates rules in the DSL into constraints enriched with 
contract-specific information. These 
constraints specify the conditions 
under which the corresponding rules 
are violated, and are then examined by the symbolic execution engine.

The core innovation lies in the design of a suite of constraint variables that 
precisely capture the diverse semantics of ERC rules. 
Some variables correspond directly 
concrete contract variables, 
while others represent abstract contract states that have no explicit source code counterparts.
We further develop program analysis mechanisms that identify the former when translating DSL rules into contract-specific constraints, and define how to update the latter
during symbolic execution.
Using state variables simplifies the solver’s input compared to using constraint predicates or functions to represent contract states, 
as state variables' values are determined during symbolic execution.
Together, these capabilities enable \Tool{} to handle the variation across various contract implementations, distinguishing it from existing symbolic execution techniques~\cite{klee,S2E,SymCC}.


\bolditalicparagraph{Constraint Variables.}
For each public function $\var{fun}$, we define six types of \emph{state variables} 
to track whether $\var{fun}$ or any of its callees perform specific actions, 
thereby transitioning into the corresponding states. 
These variables play a crucial role in representing ERC violation 
constraints. The symbolic execution 
engine initializes all state variables 
to \texttt{false} and updates them to 
\texttt{true} once the actions are conducted by $\var{fun}$.

\begin{itemize}[noitemsep, topsep=0pt, leftmargin=.25in]

\item $\var{TH}$ denotes whether $\var{fun}$ throws an exception. We consider exceptions raised by \texttt{require}, \texttt{revert}, \texttt{assert}, and \texttt{throw}. 
Exceptions triggered inside modifiers are naturally supported 
via inter-procedural analysis. 

\item $\var{EM}_{e}$ denotes whether $\var{fun}$ emits event $e$.

\item $CA_{\var{fun2}}$ denotes whether $\var{fun}$ invokes function $\var{fun2}$.

\item $\var{BI}_\var{f}$, $\var{BD}_\var{f}$ and $\var{BC}_\var{f}$ denote whether $\var{fun}$ increases, decreases, or  modifies field $\var{f}$, respectively.
\end{itemize}

Moreover, we define an $O$ variable for each instruction, 
representing the instruction sequence along an execution path. 
We define value variables for contract fields, formal and actual function parameters, event parameters, local variables, and values defined by Ethereum (\eg, \texttt{msg.sender}).

\begin{figure}
  \begin{subfigure}[t]{0.49\textwidth}
    \centering
    \scriptsize
\begin{tcolorbox}[colframe=black, boxrule=0.5pt, colback=white,
width=\textwidth,
arc=1mm, left=-1.5mm, right=0.1mm, top=-1.8mm, bottom=0.1mm]
    \begin{align}
& \textcolor{orange}{if \quad msg.sender \quad != }\quad ticketToOwner[\_tokenId]  \\
& \textcolor{orange}{and \quad msg.sender \quad != }\quad ticketApprovals[\_tokenId] \\
& \textcolor{orange}{and \quad not} \quad \_operatorApprovals[\_from][msg.sender] \\
& \textcolor{orange}{then} \quad TH = true
\end{align}
\end{tcolorbox}
    
    \mysmallcaption{fig:gen-1}{Step 1}{}
  \end{subfigure}\hfill
  \begin{subfigure}[t]{0.45\textwidth}
  \centering
  \scriptsize
  \begin{tcolorbox}[colframe=black, boxrule=0.5pt, colback=white,
width=\textwidth,
arc=1mm, left=-1.5mm, right=0.1mm, top=-1.8mm, bottom=0.1mm]
\setcounter{equation}{0}
        \begin{align}
& \textcolor{orange}{if} \quad msg.sender \neq ticketToOwner[\_tokenId] \quad \land  \\
& msg.sender \neq ticketApprovals[\_tokenId] \quad \land \\
& \neg \_operatorApprovals[\_from][msg.sender] \\
& \textcolor{orange}{then} \quad TH = true
\end{align}
\end{tcolorbox}
    
    \mysmallcaption{fig:gen-2}{Step 2}{}
  \end{subfigure}\hfill
  
  \begin{subfigure}[t]{0.38\textwidth}
    \centering
    \scriptsize
      \begin{tcolorbox}[colframe=black, boxrule=0.5pt, colback=white,
width=\textwidth,
arc=1mm, left=-1.5mm, right=0.1mm, top=-1.8mm, bottom=0.1mm]
\setcounter{equation}{0}
    \begin{align} 
    &(\var{msg.sender} \neq \var{ticketToOwner}[\_\var{tokenId}])\ \land     \\
    & (\var{msg.sender} \neq \var{ticketApprovals}[\_\var{tokenId}]) \ \land  \\
    & \neg \_\var{operatorApprovals}[\_\var{from}][msg.sender]  \\
    &\land \; \neg (\var{TH} = \var{true}) 
    \end{align}
    \end{tcolorbox}
  \mysmallcaption{fig:gen-3}{Step 3}{}
  \end{subfigure}
  
\vspace{-0.1in}

  \mycaption{fig:gen}{Results of the three Steps in constraint generation.}{(DSL elements are shown in orange, and constraint elements are shown in black.)}

\end{figure}

\bolditalicparagraph{Generating Constraints.}
We generate constraints in three steps. In the first two steps, 
DSL elements are replaced by constraint elements, while in the final step, 
the composition structures are translated. 
Figure~\ref{fig:gen} illustrates the results of these three steps when translating 
Figure~\ref{fig:721-high-grammar}.

First, we perform static analysis on each contract and recursively translate all utility functions from the innermost to the outermost level, following the rules summarized in Table~\ref{tab:step1}.
Using Figure~\ref{fig:721-high-grammar} as an example, we replace \texttt{getField(`ownerOf()')}  with  \texttt{ticketToOwner}, since this field is the one returned by \texttt{ownerOf()} (see line 28 in Figure~\ref{fig:721-high}). 
Similarly, \texttt{getField(`getApproved()')} and \texttt{getField(`isApprovedForAll()')}   are replaced with 
\texttt{ticketApprovals} and \texttt{\_operator\discretionary{-}{}{} Approvals}, respectively.
We then replace all three \texttt{getPara()} invocations with the constraint variables representing the formal parameters of function \texttt{transferFrom()}.
Next, we substitute each \texttt{getFieldValue()} call with the corresponding field lookup operation, using the formal parameters and \texttt{msg.sender} as keys.
Figure~\ref{fig:gen-1} shows the result of this step.

Second, apart from \texttt{if}, \texttt{then}, and \texttt{with}, 
we replace all other DSL terminals with corresponding constraint variables, 
as well as Boolean and numerical operators for constraints.
Figure~\ref{fig:gen-2} shows the result of this step when translating Figure~\ref{fig:721-high-grammar}. 
%

\begin{table}
\centering
\footnotesize
\mycaption{tab:constraint}
{Transformation rules used in the third step for constraint generation.}{}

\begin{tabular}{|c|p{0.25\linewidth}|p{0.36\linewidth}|}
\hline
 \textbf{Non-Terminal} & {\textbf{Step 2}} & {\textbf{Step 3}} \\

\hline
\hline

\texttt{<throw>} & \texttt{if} $\Phi_{if}$ \texttt{then} $\Phi_\var{check}$ & $\Phi_{if} \land \neg \Phi_\var{check}$ \\ \hline
\texttt{<call>} & \multirow{2}{*}{\parbox{1\linewidth}{\texttt{if} $\Phi_\var{if}$ \texttt{then} $\Phi_\var{check}$ \texttt{with} $\Phi_\var{with}$}} 
& \multirow{ 2}{*}{\parbox{1\linewidth}{{${(\Phi_\var{if}~\land~\neg~\Phi_\var{check})~\lor}$${(\Phi_\var{if} \land \Phi_{check} \land \neg \Phi_\var{with})}$}}}  \\ \cline{1-1}

\texttt{<emit>} &  & \\ \hline 

 \texttt{<assign>} & \multirow{ 2}{*}{$ \Phi_\var{check}$}  & \multirow{ 2}{*}{$\neg \Phi_\var{check}$}  \\ \cline{1-1}
 \texttt{<follow>} &  &  \\ \hline
\end{tabular}
\end{table}

Third, we group existing constraints into \texttt{if}’s condition ($\Phi_\var{if}$)
(\eg, lines 1--3 in Figure~\ref{fig:gen-2}), 
the check part ($\Phi_\var{check}$) (\eg, line 4 in Figure~\ref{fig:gen-2}), 
and \texttt{with}’s condition 
($\Phi_\var{with}$), and 
eliminate \texttt{if}, \texttt{then}, and \texttt{with} using the rules in Table~\ref{tab:constraint}.
The intuition is that 
if a rule requires an action under a condition, its violation happens when
the condition is met but the action is not performed (\eg, \texttt{<throw>}), or the action is 
performed, but with a wrong parameter (\eg, \texttt{<emit>}).
For example, the rule in Figure~\ref{fig:721-high-grammar} follows the pattern ``\texttt{if} $\Phi_\var{if}$ \texttt{then} $\Phi_\var{check}$.''
According to the first row of Table~\ref{tab:constraint}, we eliminate the \texttt{if} and \texttt{then} by negating $\Phi_{check}$ and conjoining it with $\Phi_\var{if}$ using $\land$.
The resulting constraints are shown in Figure~\ref{fig:gen-3}.

For rules requiring no exceptions under certain conditions, 
we design validation constraints as ``$\neg \Phi_\var{if} \land \neg \Phi_\var{check}$.’’ 
A violation is 
reported only when the violation constraints are met, 
but the validation constraints are not met, 
ensuring the exception is indeed due to the specified conditions.
Additionally, for \texttt{<call>} and \texttt{<emit>}, 
we inspect whether the 
contracts contain the called functions or emitted events 
before synthesizing constraints.

\subsection{Static Analysis and Symbolic Execution}
\label{sec:se}


{\noindent{{\textit{\textbf{Static Analysis.}}}}}
The static analysis routines are for two purposes:
1)	verifying contracts contain the necessary functions and events and 
their declarations meet the requirements, and
2)	implementing utility functions, 
such as \texttt{getField()} and \texttt{getOrder()}.

%

{\noindent{{\textit{\textbf{Symbolic Execution.}}}}}
We use symbolic execution to analyze each rule individually.
For rules that apply to a specific public function, 
we perform symbolic execution directly on that function.
For rules that concern the entire contract, we analyze each public function of the contract separately.

Our symbolic execution process is similar to existing 
techniques~\cite{klee,S2E,SymCC}, but it differs in how we handle state variables.
All state variables are initialized to \texttt{false} and 
set to \texttt{true} once the corresponding actions are executed.
All other variables are treated as symbolic unless 
their values can be determined through static analysis.
At the end of each public function (\eg, upon \texttt{return} or an exception), we compute the conjunction of three types of constraints:
1) initial constraints imposed by Ethereum or variable types,
2) constraints enforced by the implementation (\eg, path conditions), and
3) violation constraints synthesized for the rule under verification.
If the solver (Z3) finds a satisfying solution for this conjunction, 
the rule is considered violated.
To reduce potential false positives caused by the LLM, 
we ignore rules that are violated by every contract within an ERC.
Additionally, we cap loop iterations and recursion depths at two to mitigate path explosion.

For example, when checking path 6-12-15-16-13 in Figure~\ref{fig:721-high}
against the rule in Figure~\ref{fig:gen-3},
$TH$ is computed as false, since no exception is thrown along the path. 
The constraints for ``\texttt{\_from}’’ 
include ``$\_\var{from} \geq 0$'' due to type requirements
and ``$\var{ticketToOwner}[\_\var{tokenId}] = \_\var{from}$'' from the path condition in line 8. 
Similarly, the constraints for ``\texttt{\_to}’’ include ``$\_\var{to} \geq 0$'' and ``$\_\var{to} \neq 0$''.
Contract field \texttt{ticketToOwner} is updated in line 15, resulting
its $BC$ being set to true. 
Z3 finds multiple solutions for the conjunction of these initial constraints, computed constraints, and
the violation constraint in Figure~\ref{fig:gen-3}.
One solution is 
``$\_from = 11$'', ``$\_to = 3$'', ``$\_tokenID = 26285$'', ``$msg.sender = 3$'', ``$ticketToOwner[26285] = 11$'', ``$ticketApprovals[26285] = 11$'',
and ``$\_operatorApprovals[11][3] = false$'',
indicating the rule is violated after  given those values.

\section{Evaluation}
\label{sec:eva}

\bolditalicparagraph{Implementation.}
We employ the GPT-5 model~\cite{gpt5} as the LLM in \Tool{}, 
interacting with it through OpenAI’s official APIs. 
The model conducts extensive internal reasoning before producing responses~\cite{reasoning}. 
The temperature parameter is fixed at 1.0 and cannot be adjusted. 
We set the reasoning-effort parameter to high to ensure the model’s good performance.


The remaining functionalities 
are implemented in Python, including generating prompts (based on the templates) to extract rules 
from ERC documents and translate natural language rules into the DSL, 
synthesizing constraints, and performing symbolic execution.
We implement the symbolic execution engine based on Slither~\cite{slither}, a static analysis framework for Solidity. Slither translates Solidity source code into an IR with control-flow and data-flow information to facilitate static analysis, 
but it does \emph{not} provide the symbolic execution capability.
Furthermore, we perform inter-procedural analysis for each call site by
using Slither’s API to identify the callee function, propagating 
necessary information
(\eg, constraints), and returning computed results back to the
caller after analyzing the callee.


\bolditalicparagraph{Research Questions.}
Our experiments are designed to answer the following research questions:
\begin{itemize}[noitemsep, topsep=0pt, leftmargin=.25in]

\item  \emph{Effectiveness}: Can \Tool{} accurately pinpoint ERC rule violations? 
\item  \emph{Advancement}: Does \Tool{} outperform existing auditing solutions?
\item  \emph{Rationality}: What are the benefits of each component of \Tool{}?

\item  \emph{Generality}: Can \Tool{} detect violations for ERCs beyond those 
studied in Section~\ref{sec:study}?

\end{itemize}

\bolditalicparagraph{Experimental Setting.}
%
%
All our experiments are performed on a server machine, with Intel(R) Xeon(R) Silver 4110 CPU @ 2.10GHz, 256GB RAM, and Red Hat Enterprise Linux 9.

\subsection{Effectiveness of \Tool{}}
\label{sec:large}

\begin{table}[t]
\centering
\footnotesize

\mycaption{tab:large}
{Evaluation results on the large dataset. }
{($x_{(y)}$: 
$x$ true positives, and $y$ false positives.
)
}
{
\vspace{-0.05in}
\begin{tabular}{|l||c|c|c||c|}
\hline
                 &  {\textbf{High}} & {\textbf{Medium}} & {\textbf{Low}}  & {\textbf{Total}} \\ 



\hline
\hline

{\textbf{ERC20}}     & $1353_{122}$  & $3686_{0}$  & $172_{0}$    & $5211_{122}$    \\ \hline

{\textbf{ERC721}}    & $18_{0}$        & $22_{0}$    & $502_{0}$   & $542_{0}$  \\ \hline
{\textbf{ERC1155}}   & $4_{0}$         & $12_{0}$    & $14_{0}$    & $30_{0}$ \\ \hline \hline
{\textbf{Total}}     & $1375_{122}$  & $3720_{0}$  & $688_{0}$  & $5783_{122}$ \\ \hline

\end{tabular}
}

\end{table}



\subsubsection{Methodology}
We create a \emph{large} dataset of 4,000 \emph{unique}
contracts for evaluation, 
including 
3,400 ERC20 contracts, 500 ERC721 contracts, 
and 100 ERC1155 contracts.
The contracts are randomly sampled from 
etherscan.io~\cite{etherscan} and polygonscan.com~\cite{polygonscan},
with the former chosen because of its prominence and the latter selected 
due to its higher contract deployment compared to 
other platforms (\eg, Arbitrum~\cite{arbitrum}, BscScan~\cite{bsc}). 
We discard contracts that lack source code or cannot be compiled during the sampling.
The collected contracts were deployed between September 2023 
and July 2024. This dataset serves as a random sample of real-world contracts
implementing the three ERCs, with enough contracts 
to support statistically confident conclusions.
On average, each contract contains 469.3 lines of source code,
with the largest one reaching 7,139 lines and 916 contracts exceeding 1,000 lines. 
The contracts are comparable in code size to 
other types of contract (\eg, DeFi),
and are suitable for evaluating \Tool{}'s scalability.

We run \Tool{} on the dataset and count the violations and 
false positives reported by \Tool{} to assess its effectiveness. 
For each flagged violation, we manually review the result of symbolic execution, 
the rule description, and the relevant contract code to determine its correctness. 
Each reported violation is reviewed by at least two paper authors, 
with an agreement rate over 94\%.
Due to the dataset’s size, we do not manually analyze all the contracts to identify 
ERC rule violations in them.
Instead, we sample the top 50 contracts 
with the highest number of detected violations, 
as they are more likely to be of lower quality and to contain more violations. Following the ERC standards, 
we manually identify all violations in these contracts and use them to count the false negatives of \Tool{}.



\subsubsection{Experimental Results}
\label{sec:effect_result}
As shown in Table~\ref{tab:large}, \Tool{} detects 5,783 ERC rule violations while 
reporting only 122 false positives, 
highlighting \emph{\Tool{}’s effectiveness and accuracy in checking ERC compliance}. Furthermore, \Tool{} captures all violations in the 50 contracts 
that are manually examined by the authors, 
showcasing \emph{its good coverage of ERC violations in real-world scenarios}.
In total, \Tool{} issues 234 GPT-5 queries in this experiment to extract rules from the three ERCs and translate them into the IR, with a total cost of \$5.49. 

{
\begin{figure}[t]

\begin{minipage}{\columnwidth}
\begin{center}
\footnotesize
\begin{lstlisting}[numbers=left, xleftmargin=.20in, framexrightmargin=-0.05in,framexleftmargin=.16in,language=Solidity,basicstyle=\ttfamily,morekeywords={-},morekeywords={+},keepspaces=true]
 contract MyERC1155Token is ERC1155 {
   mapping(uint256=>mapping(address=>uint256)) _balances;
   mapping(address=>mapping(address=>bool)) _approvals;
   
   function safeTransferFrom(address from, address to, uint256 id, uint256 value, bytes memory data) public {
+    require(to != address(0), "invalid receiver");
+    require(_approvals[from][msg.sender], "the caller is not approved to manage the tokens");
     _update(from, to, id, value);
+    if (to.code.length > 0) {
+      try IERC1155Receiver(to).onERC1155Received(msg.sender, from, id, value, data) returns (bytes4 response) {
+         ...
+      }}
   }
   function _update(address from, address to, uint256 id, uint256 v) internal {
     if (from != address(0)) {_balances[id][from] -= v;}
     if (to != address(0)) {_balances[id][to] += v;}
   }}
\end{lstlisting}
\vspace{-0.05in}
\mycaption{fig:1155-high}{An ERC1155 contract contains four high-security
impact violations.}
{(Code simplified for illustration.)}
\end{center}
\end{minipage}
\vspace{-0.25in}
\end{figure}
}

\indentitalicparagraph{True Positives.}
Among the 5,783 detected ERC rule violations, 
1,375 have a high-security impact, 3,720 have a medium-security impact, 
and 688 have a low-security impact. These violations span all three ERC standards, 
with 5,211 violations for ERC20, 542 for ERC721, and 30 for ERC1155. These results demonstrate \emph{\Tool{}’s capability to handle rules with varying security impacts across different ERC standards}.

Among the 1,353 high-security violations in ERC20 contracts, 
14 are due to not checking whether the message caller of \texttt{transferFrom(address \_from, address \_to, uint256 \_value)}
has sufficient privileges allowed by \texttt{\_from} to transfer \texttt{\_value} tokens.  
Another 1,266 cases violate the same rule in a different way. 
Instead of comparing the allowed amount with \texttt{\_value}, 
they check if it equals \texttt{type(uint256).max}.
Furthermore, they do not reduce the allowed amount after the transfer, which 
creates a backdoor
that allows an address to drain any 
amount of tokens from another account.
The remaining 73 violations breach the rule that \texttt{transfer(address \_to, uint256 \_value)} should revert if the caller 
lacks sufficient tokens. 
The flawed contracts either use
\texttt{unchecked} to 
bypass the underflow check when reducing the caller's balance 
or include a path 
failing to decrease the caller's balance, 
potentially allowing an address
to spend more tokens than it owns.

\Tool{} identifies 18 high-security violations that do not comply with ERC721 requirements. 
Exploiting these 
vulnerabilities could lead to NFTs being stolen or 
permanently lost at the zero address.
Figure~\ref{fig:721-high} shows one of those violations. 

\Tool{} pinpoints four high-security violations in an ERC1155 contract.
The simplified contract code is shown in Figure~\ref{fig:1155-high}.
Function \texttt{safeTransferFrom()} in lines 5--13 
transfers \texttt{value} tokens of type 
\texttt{id} between addresses.
Like ERC721, ERC1155 requires the message caller 
to be approved to manage the owner’s tokens. 
Unfortunately, neither \texttt{safeTransferFrom()} 
nor its callee \texttt{\_update()} performs this crucial 
check (via \texttt{\_approvals}), allowing unauthorized transfers.
The patch in line 7 adds a necessary check to ensure the message caller has
the owner’s approval, aligning with ERC1155.
Moreover, ERC1155 mandates \texttt{safeTransferFrom()} checks whether 
the recipient is a contract and, if so, ensures it can handle ERC1155 tokens by calling 
\texttt{onERC1155Received()} on it.
Unfortunately, Figure~\ref{fig:1155-high} neglects these
requirements, potentially trapping tokens in contracts without the token-handling capability. 
Lines 9--12 show the necessary checks.
The remaining violation arises from failing to check 
if the recipient is address zero. 
Transferring tokens to address zero leads to irreversible token loss.
The violation is fixed by line 6.


The 3,720 medium-security violations stem from several reasons. 
Of these, 3,611 involve ERC20 rules, 
where \texttt{transfer()} and \texttt{transferFrom()} incorrectly 
throw an exception when transferring zero tokens,
instead of treating it as a normal transfer. Additionally, 55 violations result from missing functions required 
by the ERC. 
Another 32 violations occur due to missing return values or incorrect return-value types. Finally, 22 violations stem from failing to throw an exception for invalid input parameters.

{
\begin{figure}[t]

\begin{minipage}{\columnwidth}
\begin{center}
\footnotesize
\begin{lstlisting}[numbers=left, xleftmargin=.20in, framexleftmargin=.16in, framexrightmargin=-0.05in,language=Solidity,basicstyle=\ttfamily,morekeywords={-},morekeywords={+},keepspaces=true]
 contract MintPassGELO is ERC1155 {
  address private _owner;
  mapping(uint256=>mapping(address=>uint256)) _balances;
  
  function airDrop(address frm, address[] memory to, uint id, uint cnt) public{ 
   require(_owner == msg.sender);
   for(uint256 j = 0; j < to.length; j++) {
    _mint(frm, to[j], id, cnt);
   }
  }
  function _mint(address from, address to, uint256 id, uint256 amount) internal virtual {
   _balances[id][to] += amount;
-  emit TransferSingle(msg.sender, from, to, id, amount);
+  emit TransferSingle(msg.sender, address(0), to, id, amount);
  }}
\end{lstlisting}
\vspace{-0.05in}
\mycaption{fig:emit}{An ERC1155 contract failing to emit an event with correct parameters.}
{(Code simplified for illustration.)}
\end{center}
\end{minipage}
\vspace{-0.25in}
\end{figure}
}


Of the 688 event-related violations, 672 are due to missing event declarations, 
15 fail to emit an event, and one emits an event but with a wrong parameter.
Figure~\ref{fig:emit} shows the last case.
According to ERC1155, a \texttt{TransferSingle(address indexed \_operator, address indexed \_from, address indexed \_to, uint256 \_id, uint256 \_value)} event must be emitted whenever tokens are transferred.
If the transfer is to mint new tokens, formal argument \texttt{\_from} must be set to zero.
In Figure~\ref{fig:emit}, function \texttt{airdrop()} increases \texttt{\_balances} without decreasing it, indicating tokens are minted.
Although line 13 emits a \texttt{TransferSingle} event, argument \texttt{\_from} is incorrectly set to variable \texttt{from} instead of zero, thereby violating the ERC1155 rule.


\indentitalicparagraph{False Positives.}
\emph{\Tool{} demonstrates high accuracy}, reporting 122 false positives across the 
4,000 contracts. 
The false discovery rate is 2.0\%.
The false positives are due to three reasons.
First, 116 occur because the actions required by ERCs are in external contracts, and their code is unavailable during evaluation.
Second, five false positives are due to \Tool{}'s inability to handle assembly code.
Third, the remaining false positive arises from an implementation of \texttt{transferFrom()} of an ERC20 contract, 
where the source address is compared with the message caller and an authorization check is only performed if they differ. 
We consider this a false positive, as the authorization check is skipped only when 
the caller is transferring her own tokens, 
making exploitation impossible.
Notably, the ERC20 standard does not require implementations to check the 
caller’s authorization only when it differs from the token owner. 
This flexibility allows scenarios in which a token owner restricts how 
many of her own tokens she can spend (\eg, to prevent overspending).

\indentitalicparagraph{False Negatives.}
The 50 contracts with the highest number of detected violations consist of 42 ERC20 contracts, five ERC721 contracts (including Figure~\ref{fig:721-high}), 
and three ERC1155 contracts (including Figure~\ref{fig:1155-high}). 
Two authors independently examine these contracts against the corresponding ERC standards and identify a total of 254 violations. 
Among these, 48 have a high-security impact, 158 have a medium-security impact, 
and 48 have a low-security impact. In total, the contracts 
violate 38 distinct rules, including 15 ERC20 rules, 
9 ERC721 rules, and 14 ERC1155 rules.

Comparing the manually identified violations with those detected by \Tool{}, 
we find that \Tool{} successfully detects all violations present in these contracts. 
While \Tool{} is unable to detect certain classes of violations (\eg, violations related to return-value generation) due to design and implementation limitations, 
such cases are rare in practice. Overall, \emph{\Tool{} provides strong 
coverage of real-world violations in ERC contracts}.

\indentitalicparagraph{Results of the LLM.}
We manually inspect the LLM’s outputs to assess their 
impact on \Tool{}’s results, which takes less than one day. 
Note that \Tool{} itself is fully automated and does not 
require any manual review or correction of LLM outputs during normal use.

For ERC rule extraction, the LLM successfully identifies all 132 rules across the three ERC documents.
Additionally, it mistakenly extracts six non-existent rules, 
five incorrectly require functions not to throw exceptions under certain conditions, and one pertains to ERC20’s \texttt{approve()} function, requiring the function to set allowance to zero before reassignment.
The latter requirement, however, applies to Ethereum client code that interacts with \texttt{approve()}, rather than to the implementation of the function itself.
Since these rules are violated by all contracts of the ERCs, 
\Tool{} automatically ignores them (see Section~\ref{sec:se}).

Out of the 138 extracted rules (including the six incorrect ones), 
42 are function or event interfaces that do not require translation. Among the remaining 96 rules, 
the LLM successfully translates 69 but fails on 27 for two reasons. 
First, 24 rules govern how to generate return values, but we do not define any DSL 
for return-value rules
(see Section~\ref{sec:translation}). 
As a result, \Tool{} skips their translation. Second, three rules lack sufficient detail in their textual descriptions, making the LLM not generate anything. 
For instance, ERC1155 requires \texttt{safeTransferFrom()} to throw on any error, 
but it does not specify all possible errors.

\indentboldunderparagraph{Answer to effectiveness:}
{\it{
\Tool{} accurately detects various types of ERC rule violations, 
many clearly linked to potential financial losses.
}}

\subsection{Comparison with Baselines}
\label{sec:compare}


\subsubsection{Methodology}
Since the large dataset lacks ``\emph{ground-truth}'' 
labels, we 
create a small, labeled dataset to compare \Tool{} with existing auditing solutions.
We randomly select ERC20 contracts 
audited by the Ethereum Commonwealth Security 
Department (ECSD), an expert group that reviews GitHub-submitted audit requests
and publishes their audit results on GitHub~\cite{humanaudited}. 
The selection criteria are: 
1)	Solidity source code provided, 2) approval by Solidity programmers 
with the ``approved'' tag, 
3) ERC rule violations identified, 
and 4) all code in a single contract file.
These contracts and the audit results allow us to compare 
automated tools with expert reviews. 
As ECSD has limited ERC721 and ERC1155 contracts,
we supplement our dataset with samples 
from ERCx~\cite{ERCx}.
In total, the ground-truth dataset consists of 40 contracts: 
30 ERC20, five ERC721, and five ERC1155, with an average of 553 lines of code per contract.
These contracts were either audited by ECSD between January 2019 and October 2023,
or analyzed by ERCx between May 2024 and December 2024.
After careful inspection, we identify 159 violations: 
28 high-impact, 55 medium-impact, and 76 low-impact.

%

We compare \Tool{} against
five static analysis tools, including Slither’s ERC-related checkers\footnote{arbitrary-send-erc20, erc20-
interface, erc721-interface, arbitrary-send-eth, arbitrary-send-erc20-permit, and slither-check-erc},
AChecker~\cite{ghaleb2023achecker},
ZepScope~\cite{ZepScope}, 
NFTGuard~\cite{yang2023definition}, 
and Mythril~\cite{mythril},
and a dynamic analysis tool, \textit{ERCx}~\cite{ERCx}.
Other recent techniques either rely on manually 
written, contract-specific rules~\cite{li2020securing,certora,8952204}, require access tokens~\cite{VerX}, lack publicly available source code~\cite{liu2025demy},
or depend on outdated Etherscan APIs~\cite{liu2022invcon, liu2022finding}, 
making direct comparison infeasible. 
These six tools represent the state of the art in ERC compliance validation. 
We also compare \Tool{} with ECSD, 
as ECSD is the only human auditing service whose audit results are available to us.

\begin{table}[t]
\centering
\footnotesize

\mycaption{tab:compare}
{Evaluation results on the ground-truth dataset.}
{($x_{(y, z)}$: x true positives, y false positives, and
z false negatives.)
}
{
\begin{tabular}{|l|c|c|c|c|c|c||c|c|}
\hline
    & {\textbf{Slither}} & {\textbf{AChecker}} & {\textbf{ZepScope}} & {\textbf{NFTGuard}} & {\textbf{Mythril}} & {\textbf{ERCx}} & {\textbf{\Tool{}}}  \\ 

\hline
\hline

{\textbf{High}}  & $0_{(0, 28)}$ & $0_{(0, 28)}$ & $0_{(0, 28)}$ & $0_{(0, 28)}$  & $0_{(0, 28)}$ & $0_{(2, 28)}$ &  $28_{(1, 0)}$   \\ \hline
{\textbf{Medium}}  & $26_{(0, 29)}$ & $0_{(0, 55)}$ & $2_{(19, 53)}$  & $ 0_{(0, 55)}$ & $0_{(0, 55)}$ & $12_{(21, 43)}$  & $53_{(0, 2)}$    \\ \hline
{\textbf{Low}}  & $13_{(0, 63)}$ & $0_{(0, 76)}$ & $0_{(0, 76)}$ & $0_{(0, 76)}$  & $0_{(0, 76)}$  & $0_{(0, 76)}$ & $76_{(0, 0)}$  \\ \hline  
{\textbf{Total}}  & $39_{(0, 120)}$  & $0_{(0,159)}$ & $2_{(47,157)}$  & $0_{(0, 159)}$ &  $0_{(0, 159)}$ & $12_{(23,147)}$ & $157_{ (1, 2)}$   \\ \hline

\end{tabular}
}

\vspace{-0.1in}
\end{table}


\subsubsection{Experimental Results}

As shown in Table~\ref{tab:compare}, \Tool{} detects 157 out of the 159 violations, 
\emph{far outperforming the six baseline techniques,} with only one false positive. 
Most detected bugs and the false positive share code patterns observed in 
the large dataset. 
Since \Tool{} does not check rules on return value generation, 
it misses two violations. Nonetheless, 
it detects all others, \textit{demonstrating strong coverage of ERC rule violations.}

\indentitalicparagraph{Slither} detects the highest number 
of violations among all baselines, including 26 
cases where an ERC-mandated function is either missing or does not match the required API, and 13 event-related cases.
Some of its checkers aim to detect errors in 
verifying the message caller’s privileges. However, these checkers mainly target contracts that interact with ERC-compliant contracts rather than the ERC-compliant contracts themselves. 
Consequently, the buggy code patterns they cover are absent from the 
ground-truth dataset, and they fail to capture any of the high-impact 
violations in the ground-truth dataset.

\indentitalicparagraph{AChecker} targets access control vulnerabilities 
characterized by code patterns where critical instructions lack protection 
or modifications to contract fields containing access control information 
are unguarded. However, all access-control-related violations in the ground-truth dataset are 
cases where access control checks are present but they fail to comply with ERC requirements. Consequently, AChecker fails to identify any of these violations.

\indentitalicparagraph{ZepScope}
reports 49 missing-check cases based on OpenZeppelin’s code. Only two are real bugs: in an ERC721 contract, \texttt{ownerOf()} and \texttt{balanceOf()} fail to check whether their inputs are zero.
These two are also detected by \Tool{}. 
The remaining are false positives: 37 where the check exists but ZepScope misidentifies it as missing, and ten where the check is absent but poses no security or functionality risk. 
For example, ZepScope flags an ERC721 contract's \texttt{approve()} for not verifying the input address differs from the message sender. Yet allowing a caller to approve herself is harmless, as she already has that privilege.
Meanwhile, ZepScope misses 157 violations, showing that rules inferred solely from OpenZeppelin’s code are inadequate for detecting ERC violations.

\indentitalicparagraph{NFTGuard} 
reports the contracts do not have five types of issues 
(\eg, reentrancy, unlimited minting),
but fails to detect any ERC rule violations. Relying on NFTGuard’s bug patterns alone is insufficient for identifying ERC compliance issues.

\indentitalicparagraph{Mythril}’s repository includes predefined constraints for detecting 11 categories of smart contract vulnerabilities. 
However, none of these constraints 
are related to ERC rules. Consequently, Mythril does not report 
any ERC rule violations when analyzing the ground-truth dataset.

\indentitalicparagraph{ERCx}
detects 12 violations: five involving ERC API non-compliance, six where implementations forbid zero 
input values despite ERC20 explicitly allowing them, and one where \texttt{getApproved(\allowbreak uint256 \_tokenID)} 
in an ERC721 contract fails to validate its input.
However, many violations are missed because ERCx’s unit tests are incomprehensive to dynamically trigger them.
ERCx reports 23 false positives. Six stem from misinterpreting ERC rules, such as flagging four APIs as non-compliant but they are 
not required by the ERCs. 
The remaining 17 result from flawed unit tests. 
For instance, ERCx reports two violations where \texttt{safeBatchTransferFrom()} does not invoke a  function required by ERC1155. 
In reality, the call is only mandatory when transferring tokens to a contract, a case not triggered by the tests.

\indentitalicparagraph{Comparison with the ECSD Auditing Service.}
We compare \Tool{} with ECSD using the 32 contracts from ECSD's GitHub. 
ECSD detects 68 violations, all of which \Tool{} successfully identifies, 
along with 70 more violations missed by ECSD. 
ECSD produces 12 false positives---two due to auditors 
not knowing Solidity auto-generates 
getter functions for contract fields, and ten from incorrectly 
assuming \texttt{transferFrom()} must verify the owner's balance, 
similar to \texttt{transfer()}, despite no such ERC20 requirement.

ECSD experts spent 128 days auditing the contracts, 
measured from the GitHub request submission to the final expert review, 
while \Tool{} completes the analysis in just 1,037 seconds, 
measured as the sum of the ChatGPT analysis time 
for the ERCs and the code analysis time for all contracts.
Additionally, ECSD charges \$160,000 for auditing (estimated by
the average hourly rate and time spent),
whereas \Tool{} costs only \$2.94 for the use of ChatGPT\footnote{ERC1155 is not involved, so the number differs from Section~\ref{sec:effect_result}.}. 
Although this is a best-effort comparison, it clearly highlights \Tool{}'s vastly lower time and monetary costs.

\indentboldunderparagraph{Answer to advancement:}
{\it{
\Tool{} detects significantly more violations than the baselines while minimizing false positives, monetary cost, and execution time, 
demonstrating its clear advantage over existing auditing solutions.
}}

\begin{figure}[t] 
    \centering
    \includegraphics[width=0.9\columnwidth]{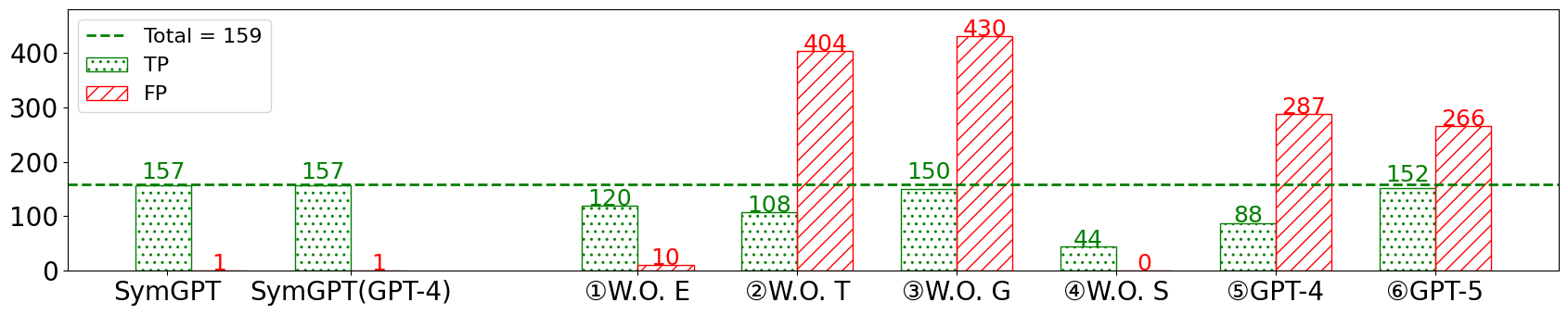} 
\mycaption{fig:ablation}{Contributions of \Tool{}'s components.}
{(There are 159 violations in the ground-truth dataset. W.O.: without, E: rule extraction, 
T: rule translation, 
G: constraint generation, and
S: symbolic execution.)}
\end{figure}

\subsection{Rationality of \Tool{}'s Components}
\subsubsection{Methodology}
To evaluate the contribution of each component in \Tool{}, 
we conduct an ablation study by disabling them individually 
and constructing multiple variants of \Tool{}. Specifically:
\textcircled{1} \textit{Without rule extraction.} We directly prompt the LLM to translate ERC documents’ specification sections into individual rules in the DSL.
\textcircled{2} \textit{Without rule translation.} We ask the LLM to generate constraints directly from the natural language rules.
\textcircled{3} \textit{Without constraint generation.} We provide the LLM with each rule in the DSL, along with the corresponding public function, its direct and indirect callees, and all contract fields accessed by the public function and all its callees (collectively referred to as the related contract code). The LLM is then asked to determine whether the rule is violated.
\textcircled{4}	\textit{Without symbolic execution.} We supply the LLM with the violation constraints and the related contract code, and request it to assess whether the constraints can be satisfied.
We compare these ablated variants with the full-featured \Tool{} on the ground-truth dataset to assess the necessity and rationale of each component.
Figures 12, 13, 14, and 15 in the appendix show the prompt templates
designed for these variants, respectively.

In addition, we evaluate a setting where the LLM analyzes each extracted natural language rule directly against the related contract code, a scenario where ERC compliance validation relies \emph{solely on the LLM}.
Besides GPT-5 (\textcircled{6} in Figure~\ref{fig:ablation}), 
we further test GPT-4 (\textcircled{5} in Figure~\ref{fig:ablation}) to examine how ERC 
compliance validation performance evolves across successive generations of LLMs. 
Figure 16 in the appendix shows the prompt template.

Since GPT-5 was released on August 7, 2025, 
and many contracts may have been used in its training, 
we additionally evaluate the performance of \Tool{} using GPT-4 as 
the underlying LLM. The corresponding results are labeled ``\Tool{} (GPT-4)’’ in Figure~\ref{fig:ablation}.

\subsubsection{Experimental Results}

As shown in Figure~\ref{fig:ablation}, the full-featured version of 
\Tool{} (labeled ``\Tool{}’’) detects more violations and 
reports fewer false positives than any of its ablated variants,  highlighting \emph{the necessity of each component}.
Replacing the underlying LLM with GPT-4 (labeled ``\Tool{} (GPT-4)'') yields the same results, demonstrating that \emph{\Tool{} is robust across 
different LLMs and does not rely on the evaluated contracts 
having appeared in its LLM’s training data}.

\textcircled{1} Without rule extraction, \Tool{} generates fewer rules in the DSL because it directly processes the entire specification section, 
leading to 39 missed violations.
\textcircled{2} Disabling rule translation causes the LLM to struggle with generating constraints that depend on contract-specific details (\eg, contract fields) and to introduce errors when synthesizing constraints. Consequently, it misses 51 violations and produces 404 false positives.
\textcircled{3} When constraint generation is disabled and the LLM is 
asked to identify violations based on rules in the DSL, 
it captures 
almost all violations in the ground-truth dataset (150/159). However, hallucinations of the LLM lead to 430 false 
positives, the highest among all configurations. 
\textcircled{4} Comparing the results of disabling 
constraint generation with those of disabling 
symbolic execution, we observe that providing the LLM with rules in constraints, rather than rules in the DSL, reduces 
both detected violations and false positives.

\textcircled{6} Relying solely on GPT-5 to check ERC rules in natural language against related 
contract code pinpoints most ERC rule violations in the dataset (152/159).
The missed violations can be categorized into three types. 
First, ERC721 requires events \texttt{Transfer}, \texttt{Approval}, and \texttt{ApprovalForAll} to be declared with \texttt{indexed} to attribute their parameters. The LLM fails to detect five declarations that violate these requirements.
Second, the LLM overlooks one instance in which the \texttt{transferFrom()} function does not verify the privilege of \texttt{msg.sender}.
Third, in one case, the \texttt{totalSupply()} function of an ERC20 contract does not return the actual total number of supplied tokens, 
but the LLM fails to identify this violation. \Tool{} also misses this issue 
because it lacks the capability to validate rules that involve generating specific return values.
On the other hand, using only GPT-5 results in 266 false positives, far higher than \Tool{}, which reports only a single false positive. This result highlights the advantage of combining symbolic execution with GPT-5.
\textcircled{5} Compared with GPT-5, using GPT-4 alone identifies 
fewer true violations while producing more false positives, 
highlighting the necessity of leveraging more recent LLMs when validating ERC compliance.

\indentboldunderparagraph{Answer to rationality:}
{\it{
Each component of \Tool{} contributes to its effectiveness, and  combining symbolic execution with the LLM significantly enhances the accuracy of ERC rule validation.
}}

\subsection{Generality of \Tool{}}
\label{sec:3525}

%




\subsubsection{Methodology}


We use ERC3525 and ERC4907 to evaluate whether \Tool{} can validate contracts beyond the ERCs we have studied against potentially more complex ERC rules. 
ERC3525 is designed for semi-fungible tokens, where each token is unique like an NFT but also possesses a quantitative nature similar to fungible tokens~\cite{erc3525}. 
ERC4907 extends ERC721 by introducing time-limited usage rights: for each NFT, it allows a designated user to use the token within a specified time window while prohibiting transfer during that period~\cite{erc4907}.
Both ERC3525 and ERC4907 have already been adopted in financial instruments~\cite{fujidao,bufferfinance,4907-usage}. Introduced at least 
two years after the ERCs analyzed in Section~\ref{sec:study}, both ERC3525 and ERC4907 
represent more recent developments in the ERC ecosystem. 
Following the same methodology, 
we manually identify 58 rules in ERC3525, 
and 12 rules in ERC4907.

We do not find any ERC3525 or ERC4907 contracts on
etherscan.io or polygonscan.com.
Therefore, we search GitHub for contracts of these ERCs. 
We find only one ERC3525 contract on GitHub. In contrast, many ERC4907 contracts are available, from which we randomly select ten.
After manual inspection, we confirm that only one ERC4907 contract contains a single violation, where function \texttt{setUser()} fails to inspect whether the input
\texttt{tokenId} is valid, as required by ERC4907, 
while all other contracts fully comply with ERC3525 or ERC4907. 
To evaluate \Tool{}, we inject violations into the compliant contracts 
by randomly removing code related to ERC compliance. 
For the ERC3525 contract, we perform random error injection ten times, injecting three errors each time. For the remaining nine ERC4907 contracts, we inject three errors into each contract. 
In total, we construct a dataset consisting of 20 contracts: 
19 contracts each contain three injected violations, 
and one contract contains a single violation originally introduced by its developer.

\subsubsection{Empirical Study Results}

We conduct the same empirical study as 
Section~\ref{sec:study} on the 70 rules in ERC3525 and ERC4907.
Regarding rule content, nine rules pertain to privilege checks, 24 relate to functionality requirements, 18 define public functions' APIs, 
and 19 specify logging requirements.
In terms of security impact, 16 rules are classified as having a high-security impact, 
35 as medium, and the 19 logging-related rules as low.
Of the 70 rules, 23 concern function or event declarations, 
which are expressed in Solidity source code. 
Among the remaining 47 rules, 42 can be matched using the linguistic patterns listed in Table~\ref{tab:linguistic}. Consistent with the 
three previously studied ERCs, most of these rules fall under four 
primary linguistic patterns: TP1, EP1, EP2, and RP1.

\subsubsection{Experimental Results}
After running \Tool{} \emph{without any modifications} on the 20 contracts, it successfully detects all 58 violations with zero false positives. 
In the intermediate steps, the LLM correctly identifies all required rules during rule extraction, but it erroneously extracts three additional rules. One rule talks about setting the return value of function \texttt{transferFrom()}, and the other two require
function \texttt{userOf()} and function \texttt{userExpires()} not to throw when returning zero. 
During rule translation, \Tool{} fails to translate these extra rules, along with 18 other rules related to return value generation. Thus, those erroneously extracted rules
do not introduce any false positives. 
\Tool{} does not make any errors in all other steps.

%

\indentboldunderparagraph{Answer to generality:}
{\it{
The findings from the empirical study and \Tool{}'s violation detection capabilities are not limited to the 
three studied ERCs but generalize to a broader range of ERC standards.
}}

\section{Limitations and Discussion}

\indentitalicparagraph{Threats to Validity.}
The validity of our study may be affected by several factors, including the limited number of ERCs and smart contracts analyzed, 
the use of the LLM, the absence of dynamic validation for the detected violations, 
and the unsoundness of \Tool{}’s static analysis.

In Section~\ref{sec:study}, we analyze three widely adopted ERCs that illustrate common development practices and typical issues in smart contracts. 
Although many ERCs remain unexplored, we regard these three as representative examples. As demonstrated in Section~\ref{sec:3525}, the findings and linguistic patterns identified in our study also generalize to ERC3525 and ERC4907, two previously unexamined standards. 
Moreover, \Tool{}, which was developed based on these insights, 
successfully detects all injected and original violations of the two ERCs.
To evaluate \Tool{}, we constructed three datasets comprising 
over 4,000 contracts collected from five different sources. 
While additional repositories such as Arbitrum~\cite{arbitrum} and BscScan~\cite{bsc} are available, we are not aware of substantial 
differences between contracts from those sources and 
the ones included in our datasets. 
Therefore, we consider our datasets to be a best-effort representation 
of real-world ERC implementations.

\Tool{} employs GPT-5 as its underlying LLM. 
Although GPT-5 represents the state of the art at the time of writing, 
the field of LLMs is advancing rapidly, 
and it may soon be surpassed by newer models. 
Furthermore, our evaluation adopts a single configuration of GPT-5 without exploring alternative parameter settings. These limitations, however, are 
mitigated by \Tool{}’s design: the DSL constrains the LLM to a structured translation task, while critical bug detection is carried out deterministically through symbolic execution. 
Together, these features reduce \Tool{}’s dependence on any specific LLM or configuration.

We evaluate \Tool{} on over 4,000 smart contracts and identify more than 5,000 ERC rule violations. 
All detected results are carefully reviewed by the authors 
to ensure their accuracy. Section~\ref{sec:eva} presents a detailed 
categorization of the detected violations and reported false positives, 
demonstrating our thorough understanding and analysis of the results.
We do not perform dynamic validation because it would require deploying each contract locally (including downloading all external dependencies), 
configuring the execution context (\eg, assigning appropriate values 
to state variables), and creating inputs to trigger the 
relevant functions. Conducting these steps at this scale is impractical. This practice aligns with many prior research papers on static bug detection.

\Tool{} contains three static-analysis components: API compliance checkers, utility functions when synthesizing constraints, and symbolic execution engine. 
The first two are sound, as their functionality is straightforward. 
However, the symbolic execution engine has several limitations that may lead to both false positives and false negatives: 
it bounds loop iterations and recursive call depth to 
two, cannot handle assembly code, and cannot analyze external calls when the code of external contracts is unavailable.
We did not observe issues caused by the first limitation; however, the latter two result in the false positives reported in Section~\ref{sec:large}.

\indentitalicparagraph{Discussion.}
Some tools support custom rules, assertions, or contract behaviors~\cite{VerX,mythril,certora,8952204,10.1145/3656416}. SymGPT could integrate with them by translating ERC rules into the required input format and extending their functionalities to capture ERC-specific semantics. 
For instance, for rule requiring functions to throw an exception when invoked with certain inputs, \Tool{} can generate specifications in Certora’s input format, by designating target functions, assuming input values, and instructing Certora to verify whether the functions throw an exception under the inputs.
%
Likewise, extending the DSL and the symbolic execution engine would allow 
\Tool{} to generate constraints from natural language descriptions beyond ERCs and detect a broader range of smart contract issues. 
For example, 
we can extend the engine by assigning a financial type to each variable
and define a DSL to specify which financial types can be summed.
The LLM could then translate natural language descriptions of 
financial-type rules into the DSL, allowing \Tool{} to detect accounting errors~\cite{zhang2024towards}.
We leave these directions for future work.

\section{Conclusion}

This paper presents an empirical study on the implementation rules in ERC documents,
focusing on their content, security implications, and detailed specifications in natural language. Based on our findings, we developed \Tool{}, 
an automated tool that integrates an LLM with symbolic execution to 
assess smart contracts’ compliance with ERC rules. 
\Tool{} effectively detects numerous rule violations 
and outperforms six automated techniques and a manual auditing service. 
This project enhances the understanding of ERC rules and their violations, 
promoting further exploration in the field. Moreover, our work demonstrates 
the advantages of combining LLMs with formal methods 
and encourages continued research in this direction.
\newpage

\section*{Data-Availability Statement}
This work provides an open-source artifact that 
supports the claims made in our paper. The 
artifact includes the empirical study results 
of ERC rules, the source code of \Tool{}, 
evaluation datasets, experimental outputs, and scripts to fully automate all experiments. The 
scripts interact with public APIs offered by OpenAI, and they requires users to have a valid OpenAI API key set up in advance. 
Apart from standard open-source licensing terms, 
the artifact imposes no additional restrictions. It is available at \url{https://github.com/symbolic-gpt/symgpt} and submitted for artifact evaluation.

\section*{Acknowledgments}

We are grateful to the anonymous reviewers for their insightful comments and suggestions. This work was supported by the UK Engineering and Physical Sciences Research Council (EPSRC) under grants EP/T006544/2, EP/T014709/2, and EP/Z533749/1; by the Horizon Europe project TaRDIS (grant agreement 101093006; UKRI number 10066667); by the U.S. National Science Foundation (NSF) under awards CNS-1955965 and CCF-2145394; and by the National Natural Science Foundation of China under grant 92582108.

\bibliographystyle{ACM-Reference-Format}
\bibliography{gpt-1}

\clearpage
\clearpage
\section*{Appendix}

\begin{table}[H]
\centering
\footnotesize


\mycaption{tab:words}
{ Words for Abbreviation}
{\textit{}
}

\begin{tabular}{|c|l|l|}
\hline
 {\textbf{Abbr}} & {\textbf{Words}} & {\textbf{POS}}\\

\hline
\hline

SUB & a function, a contract & NSUBJ, DOBJ \\ \hline
MUST  & ``must'', ``must not'', ``cannot'', ``should'' & AUX\\ \hline
THROW & `throw'', ``revert'', ``be treated as normal'' & ROOT\\ \hline 
COND & an if statement, an unless statement & ADVCL\\ \hline

ACTION & an operation performed by a function & NSUBJ, DOBJ \\ \hline
CALLER & ``caller'' & NSUBJ \\ \hline
APPROVE & ``be approved to'' & ROOT \\ \hline 
BE  & ``are considered'' & ROOT \\ \hline 
INVALID  & ``invalid'' & ADJ \\ \hline

CALL   & ``call'' & ROOT \\ \hline
VAR    & a variable & NSUBJ \\ \hline
ASSIGN & ``set to'', ``overwrite'', ``be reset'', ``allow'', ``be set'', ``be'', ``=='', ``create'' & ROOT \\ \hline
PREP & ``with'', ``to'' & ADP \\ \hline
VALUE & a specific value & DOBJ  \\ \hline

EVENT & an event & NSUBJ \\ \hline
EMIT  & ``fire'' , ``trigger'', ``emit'' & ROOT \\ \hline

RETURN & ``return'', ``@return'' & ROOT \\ \hline

FOLLOW & ``follow'' & ROOT \\ \hline
ORDER & a specific order & DOBJ  \\ \hline

\end{tabular}

\end{table}









\begin{figure}[H]
\footnotesize
\begin{tcolorbox}[
  enhanced,
  colback=promptbg,
  colframe=promptborder,
  boxrule=0.3pt,
  arc=2mm,
  left=2mm,
  right=2mm,
  top=1mm,
  bottom=1mm,
  drop shadow,
  width=1\linewidth
]
\begin{lstlisting}[
  language=GPTPrompt,
  basicstyle=\ttfamily\footnotesize,
  numbers=none,
  frame=none,
  columns=fullflexible,
  keepspaces=true,
  showstringspaces=false]
Given the ERC, return all the rules in the ERC with the given constraints.
ERC:"""
{{specification}}
"""
Constraint JSON Schema:"""
{{schema}}
"""
Return all constraints in single JSON array: [{
    "target": "string, optional, can either be function interface or event name which rule should apply, or empty if the rule should be applied to every function",
    "constraint": "constraint object, can be one of CallVerify, EmitVerify, OrderVerify, ReturnVerify, StateAssignVerify, or ThrowVerify which follows the constraint JSON schema"
}]
\end{lstlisting}
\end{tcolorbox}
\mycaption{fig:without-rule-extraction}{The prompt template for translating the rules in an ERC’s specification section into the DSL.}
{(``\{\{specification\}\}'' refers to the specification section of an ERC, 
and ``\{\{schema\}\}'' denotes the JSON schema defining the DSL.
)}
\end{figure}

\begin{figure}[h]
\footnotesize
\begin{tcolorbox}[
  enhanced,
  colback=promptbg,
  colframe=promptborder,
  boxrule=0.3pt,
  arc=2mm,
  left=2mm,
  right=2mm,
  top=1mm,
  bottom=1mm,
  drop shadow,
  width=1\linewidth
]
\begin{lstlisting}[
  language=GPTPrompt,
  basicstyle=\ttfamily\footnotesize,
  numbers=none,
  frame=none,
  columns=fullflexible,
  keepspaces=true,
  showstringspaces=false]
Given the rule and the constraint syntax, generate the buggy constraints for the given rule and code.
Constraints can be concatenated with the ==, !=, <, <=, >, >=, &, ||,!, and ().
Interface:"""
{{API}}
"""
Rule: """
{{rule}}
"""
Code:"""
{{code}}
"""
Constraint Syntax:"""
{{syntax}}
"""
Return JSON format: {"constraints": "string, the buggy constraints"}
\end{lstlisting}
  \end{tcolorbox}
\mycaption{fig:prompt4}{The prompt template for translating a rule written in natural language into constraints.}{
(``\{\{API\}\}'' denotes the declaration of the function to which the rule applies,
``\{\{rule\}\}'' represents the rule described in natural language, 
``\{\{code\}\}'' refers to the related contract code, and
``\{\{syntax\}\}'' specifies the constraint format.
)
}

\end{figure}

\begin{figure}[h]
\footnotesize
\begin{tcolorbox}[
  enhanced,
  colback=promptbg,
  colframe=promptborder,
  boxrule=0.3pt,
  arc=2mm,
  left=2mm,
  right=2mm,
  top=1mm,
  bottom=1mm,
  drop shadow,
  width=1\linewidth
]
\begin{lstlisting}[
  language=GPTPrompt,
  basicstyle=\ttfamily\footnotesize,
  numbers=none,
  frame=none,
  columns=fullflexible,
  keepspaces=true,
  showstringspaces=false]
Audit the following code against the provided rule:"""
{{rule}}
"""
Rule Schema:"""
{{schema}}
"""
Code:"""
{{code}}
"""
Return YES if violated. NO otherwise. DO NOT EXPLAIN ANYTHING ELSE.
\end{lstlisting}

\end{tcolorbox}
\mycaption{fig:prompt5}{The prompt template for checking whether the given code violates a rule in the DSL.}{
(``\{\{rule\}\}'' represents a rule in the DSL,
``\{\{schema\}\}'' denotes the JSON schema of the DSL, and
``\{\{code\}\}'' refers to the related contract code of the rule.
)}

\end{figure}

\begin{figure}[h]
\footnotesize
\begin{tcolorbox}[
  enhanced,
  colback=promptbg,
  colframe=promptborder,
  boxrule=0.3pt,
  arc=2mm,
  left=2mm,
  right=2mm,
  top=1mm,
  bottom=1mm,
  drop shadow,
  width=1\linewidth
]
\begin{lstlisting}[
  language=GPTPrompt,
  basicstyle=\ttfamily\footnotesize,
  numbers=none,
  frame=none,
  columns=fullflexible,
  keepspaces=true,
  showstringspaces=false]
Audit the following code with the initial z3 constraints and buggy constraints.
Verify whether the code satisfied the buggy constraints if the entrypoint is the function {{entryfunction}}
Initial Constraints:"""
{{initialconstraint}}
"""
Buggy Constraints:"""
{{buggyconstraint}}
"""
Code:"""
{{code}}
"""
Return in YES if satisfied, NO otherwise. DO NOT EXPLAIN ANYTHING ELSE.
\end{lstlisting}
  \end{tcolorbox}
\mycaption{fig:llm-checking-constraints}{The prompt template for checking whether the given code satisfies the specified constraints.}{
(``\{\{initialconstraint\}\}'' represents the initial constraints derived from contract code,
``\{\{buggyconstraint\}\}'' denotes the buggy constraints translated from rules in the DSL, and
``\{\{code\}\}'' refers to the relevant smart contract code.)
}

\end{figure}

\begin{figure}[h]
\footnotesize
\begin{tcolorbox}[
  enhanced,
  colback=promptbg,
  colframe=promptborder,
  boxrule=0.3pt,
  arc=2mm,
  left=2mm,
  right=2mm,
  top=1mm,
  bottom=1mm,
  drop shadow,
  width=1\linewidth
]
\begin{lstlisting}[
  language=GPTPrompt,
  basicstyle=\ttfamily\footnotesize,
  numbers=none,
  frame=none,
  columns=fullflexible,
  keepspaces=true,
  showstringspaces=false]
Given the related code for function {{API}} and the rule "{{rule}}", return YES if the code violates the rule, NO otherwise.
Code:"""
{{code}}
"""
\end{lstlisting}
\end{tcolorbox}
\mycaption{fig:llm-checking-natural-language}{The prompt template for checking whether the given code violates a natural language ERC rule.}{
(``\{\{rule\}\}'' denotes the rule written in natural language, and
``\{\{code\}\}'' represents related smart contract code.
)
}
\end{figure}

\end{document}